\documentclass[journal]{IEEEtran}

\ifCLASSINFOpdf
\usepackage[pdftex]{graphicx}
\graphicspath{{/figs/}}
\DeclareGraphicsExtensions{.pdf,.eps}
\else
\usepackage[dvips]{graphicx}
\graphicspath{{/figs/}}
\DeclareGraphicsExtensions{.eps}
\fi

\usepackage{xspace}
\usepackage{caption}
\usepackage{subcaption}
\usepackage[cmex10]{amsmath}
\usepackage{amssymb} 
\usepackage{cite}
\usepackage{bm}
\usepackage{algorithm,tabularx} 
\usepackage[noend]{algpseudocode}
\makeatletter
\newcommand{\multilines}[1]{%
	\begin{tabularx}{\dimexpr\linewidth-\ALG@thistlm}[t]{@{}X@{}}
		#1
	\end{tabularx}
}
\makeatother

\usepackage{algpseudocode}
\usepackage{array}

\usepackage{amsthm}
\usepackage{multirow}
\usepackage{color}
\usepackage{epstopdf}
\usepackage{endnotes}
\usepackage{framed}
\usepackage{cool} 
\usepackage{enumerate} 
\usepackage[subnum]{cases}

\newcommand{\norm}[1]{\left\lVert#1\right\rVert}

\newcommand{\SumNoLim}[2]{\ensuremath{\sum\nolimits_{#1}^{#2}}}

\newcommand{\defeq}{\mathrel{\mathop:}=}
\newcommand{\eqdef}{\mathrel{\mathop=}:}
 
\newcommand{\innProd}[2]{\ensuremath{\langle{#1}, {#2}\rangle}}

\newcommand{\bigP}[1]{\ensuremath{\bigl(#1\bigr)}}

\newcommand{\BigP}[1]{\ensuremath{\Bigl(#1\Bigr)}}

\newcommand{\tbf}[1]{\textbf{#1}}

\newcommand{\Cal}[1]{\mathcal{#1}} 
\newcommand{\DD}[1]{\mathbb{#1}} 

\theoremstyle{plain}
\newtheorem{theorem}{Theorem}
\newtheorem{corollary}{Corollary}
\newtheorem{assumption}{Assumption}
\newtheorem{lemma}{Lemma}

\theoremstyle{definition}

\theoremstyle{remark}

\newcommand{\Tcomp}{\ensuremath{T^{cmp}_s\xspace}}

\newcommand{\Tcomm}{\ensuremath{T^{com}_s\xspace}}

\newcommand{\Titer}{\ensuremath{T_s^{gl}\xspace}}
\newcommand{\Eiter}{\ensuremath{E_s^{gl}\xspace}}

\newcommand{\Ecomm}{\ensuremath{E_{s,n}^{com}\xspace}}
\newcommand{\Ecomp}{\ensuremath{E_{s,n}^{cmp}\xspace}}

\newcommand{\Opt}{\textsf{MS-FEDL}\xspace}
\newcommand{\SubCPUA}{\textsf{SUB2-c}\xspace}
\newcommand{\SubBWA}{\textsf{SUB3-c}\xspace}
\newcommand{\SubLearningA}{\textsf{SUB1-c}\xspace}
\newcommand{\SubCPUB}{\textsf{SUB2-d}\xspace}
\newcommand{\SubBWB}{\textsf{SUB3-d}\xspace}
\newcommand{\SubLearningB}{\textsf{SUB1-d}\xspace}

\newcommand{\FL}{\textsf{MS-FEDL}\xspace}
\newcommand{\FEDL}{\textsf{FEDL}\xspace}

\newcommand{\SumN}{\ensuremath{\SumNoLim{n=1}{N}\xspace}}

\begin{document}
	
	\title{Toward Multiple Federated Learning Services Resource Sharing in Mobile Edge Networks}

	\author{
	\IEEEauthorblockN{Minh N. H. Nguyen\IEEEauthorrefmark{1}, \;\;\; Nguyen H. Tran\IEEEauthorrefmark{2}, \;\;\; Yan Kyaw Tun\IEEEauthorrefmark{1}, \;\;\; Zhu Han\IEEEauthorrefmark{3}, \;\;\; Choong Seon Hong\IEEEauthorrefmark{1}}
	
	\IEEEauthorblockA{\IEEEauthorrefmark{1}Department of Computer Science and Engineering, Kyung Hee University, South Korea} 
	
	\IEEEauthorblockA{\IEEEauthorrefmark{2}School of Computer Science, The University of Sydney, Sydney, NSW 2006, Australia} 
	
	\IEEEauthorblockA{\IEEEauthorrefmark{3}Department of Electrical and Computer Engineering, University of Houston, Houston, TX 77004-4005, USA}

	}
	
	
	\maketitle
	
	\begin{abstract}
	   Federated Learning is a new learning scheme for collaborative training a shared prediction model while keeping data locally on participating devices. 
   	   In this paper, we study a new model of multiple federated learning services at the multi-access edge computing server. Accordingly, the sharing of CPU resources among learning services at each mobile device for the local training process and allocating communication resources among mobile devices for exchanging learning information must be considered. Furthermore, the convergence performance of different learning services depends on the hyper-learning rate parameter that needs to be precisely decided.
       Towards this end, we propose a joint resource optimization and hyper-learning rate control problem, namely \FL, regarding the energy consumption of mobile devices and overall learning time. We design a centralized algorithm based on the block coordinate descent method and a decentralized JP-miADMM algorithm for solving the \FL problem. Different from the centralized approach, the decentralized approach requires many iterations to obtain but it allows each learning service to independently manage the local resource and learning process without revealing the learning service information. Our simulation results demonstrate the convergence performance of our proposed algorithms and the superior performance of our proposed algorithms compared to the heuristic strategy. 
		
	\end{abstract}
	
	\IEEEpeerreviewmaketitle
	\vspace{-0.1pt}
	
	\section{Introduction} \label{S:Intro}
	Nowadays, following the great success of Machine Learning (ML) and Artificial Intelligence (AI) applications, there are more and more intelligent services that have transformed our lives. This progress has been drastically elevated by the ubiquity of device-generated data that is available to the service operator and stronger computing power at cloud data centers and mobile devices. Recently, the deployment of MEC servers at the edge networks has been acknowledged as one of the key pillars to revolutionize mobile communication by assisting cellular base stations with low latency computing capability. When compared to cloud datacenter, the machine learning training process can be done at the mobile edge network with the help of multi-access edge computing (MEC) servers, resulting in lower communication latency for exchanging learning information. Therefore, these enablers unlock the full potential of edge ML applications for the vision of truly intelligent next-generation communication systems in 6G \cite{edgeAI}.
	However, the ML applications raise a privacy concern in the data collection for training purposes. In many ML applications (e.g., Crowdtracker \cite{crowdTracker2018}, Waze Carpool \cite{FreeCommunitybasedGPS}, etc.), users are required to share their sensitive personal information (i.e., user location, user identity, user photos, etc.) to the server. Furthermore, uploading a massive amount of data throughout radio access links or the Internet to the cloud data centers is costly. Hence, the strong computation capabilities of the increasingly powerful mobile devices empower the local inference and fine-tuning of Deep Neural Networks (DNNs) model on device without sending their training data to the server \cite{DeepyEye2017,Fang2018}. On the other hand, using solely the personalized local data could lead to the overfitting problem of the local training models. Thus, sharing the local learning model parameters among user equipments (UEs) equip to build up a generalized global model is the primary idea of a brand-new ML scheme, namely federated learning \cite{mcmahanCommunicationEfficientLearningDeep2017, FederatedLearningCollaborative,  WeAreMaking2017}. The deployment of this learning scheme at the edge networks brings up latency and transmission cost reduction by sending the weight parameters of local learning models to the MEC server instead of sending device-generated data to the cloud and enhances the user privacy compared to the conventional centralized training \cite{FLinMEC2019}. Inevitably, the federated learning scheme is one of the vital enablers to bring edge intelligence into reality. 
	
	\begin{figure}[t]
		\centering
		\includegraphics[width=1.\linewidth]{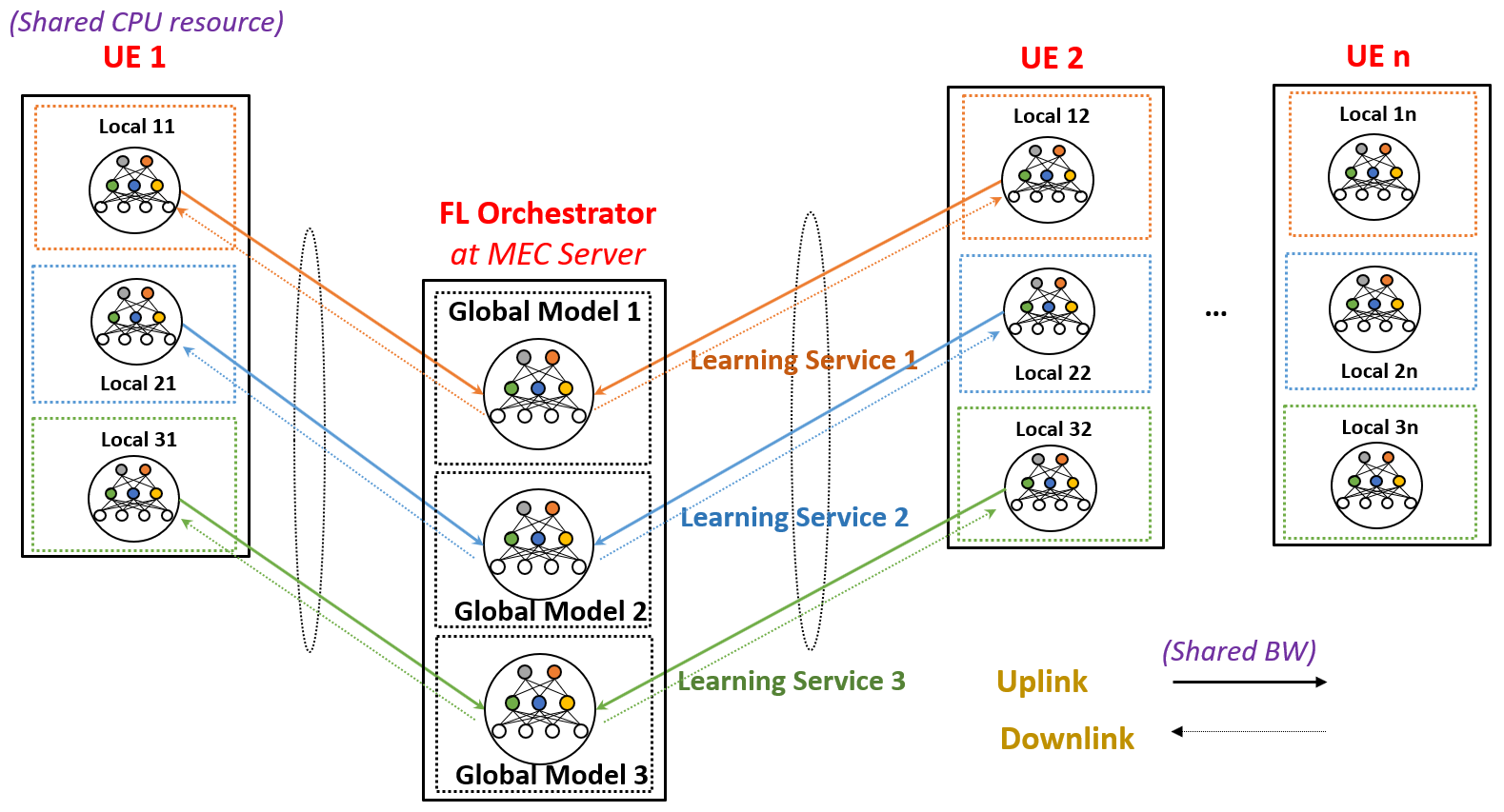}
		\caption{Multiple Federated Learning services model.}
		\label{F:System_model}
	\end{figure}
    
	In the typical federated learning scheme, the actual training process is decentralized as each UE constructs a local model based on its local dataset. Then a shared global model is updated at the server by aggregating local learning weight parameters from all UEs such as gradient, and learning weight parameters. After that the updated global model is broadcast back to UEs. 
	As an example, the work of \cite{mcmahanCommunicationEfficientLearningDeep2017} has provided the simplest form of a federated learning algorithm in which the learning parameters of the global model are averaged from local ones at UEs.  
	However, the performance of this learning scheme at the edge networks firmly depends on the allocated computation resources for local training and communication resources for uploading the updated local model parameters to the MEC server. Therefore, the allocation problem for both computation and communication resources is crucial for deploying a federated learning scheme at the edge networks. This type of problem is well-studied in \cite{tranFederatedLearningWireless2019,wangWhenEdgeMeets} that design resource allocation frameworks regarding learning performance and resource cost. Since these works focus on the model of a single federated learning service in the edge networks, there is the necessity of an extensive study and analysis for the upcoming multiple learning services systems. 
	Indeed, more data consisting of network-level mobile data (e.g., call detail record, radio information, performance indicator, etc.) and app-level data (e.g., device profile, sensing data, photos, video streaming, voice data, etc.) \cite{onDevLearning2019} can be collected on user equipment such as mobile devices, wearable devices, augmented reality headset, IoT devices. As a result, it enables many ML applications and services can be deployed on UEs or in the edge networks. In addition to the independent deployment for different learning tasks, multiple deep learning models can be deployed together and provide better performance systems such as the mobile vision system \cite{DeepyEye2017,Fang2018}. 
	  
	According to the essential deployment of multiple federated learning services, the computation resources necessarily are shared among these services for the local training while the communication resources are shared among mobile devices for exchanging learning information from different services. Moreover, the performance of learning services depends on the learning parameters that need to be precisely decided regarding the resource allocation cost and the overall learning time. In this paper, we study the under-explored problem - \emph{the shared computation, communication resource allocation, and the learning parameter control for multiple federated learning services coexisting at the edge networks}. 
	
	In Fig. \ref{F:System_model}, we depict the system of multiple federate learning services where a Federated Learning Orchestrator (FLO) at the MEC server is in charge of the computation and communication resource management, controls the hyper-learning rate parameter of learning services and operates these federated learning services. Accordingly, FLO performs two main processes as follows
	
	\textbf{Resource allocation process:} 
	In this work, we consider the flexible CPU sharing model such as the CPU frequency sharing among virtual machines or containers to perform the local learning updates. Since those virtual instances often require a high deployment cost, we consider the pre-allocating CPU strategy for different services.
	To capture the trade-off between the energy consumption of mobile devices and overall learning time, we propose a resource optimization problem, namely \FL that decides the optimal CPU frequency for each learning service and the fraction of total uplink bandwidth for each UE. As shown in Fig. \ref{F:System_model}, computing (i.e., CPU cycles) and communication (i.e., bandwidth) resources are shared among learning services and UEs, respectively. In addition to resource allocation, it also controls the hyper-learning rate of learning services such as the relative accuracy of the local learning problem at the UEs.
	
	\textbf{Federated learning process:} 
	After the shared computation and communication resources allocation, FLO performs the federated learning process iteratively according to the following steps: training local model, transmitting local learning model to FLO, updating the global model, and broadcasting the global model to UEs until the convergence is observed. 
	
	In order to provide an efficient approach for the resource allocation and learning parameter control of multiple federated learning services, in this work, we develop the problem design and analysis for FLO, which can be summarized as follows
	\begin{itemize}
		\item In Section \ref{S:MSFL_Model}, we first propose a resource-sharing model for multiple learning services. Then, we pose the resource allocation and learning parameter control problem for FLO to manage multiple learning services with the \FL problem, which is in the form of a multi-convex problem.
		\item In Section \ref{S:Sols}, we develop both centralized and decentralized approaches to solve the \FL problem. Specifically, We first propose a centralized algorithm based on block coordinate descent framework that provides a quick scheme by decoupling the global problem into three subproblems such as CPU allocation, bandwidth allocation, and hyper-learning rate decision subproblems. After that, we develop a decentralized approach based on the JP-miADMM algorithm which allows each learning service to independently manage the resource allocation, local learning process and cooperatively operate under the management of FLO. Although the decentralized approach requires many iterations to convergence, it provides a more flexible and scalable method for resource allocation without revealing the learning service information.
		\item In Section \ref{S:Sim}, we provide extensive numerical results to demonstrate the convergence performance of the proposed algorithms. Moreover, we present the performance gain of our proposed algorithms when compared with the heuristic strategy. Finally, we present the conclusions in Section \ref{S:Concls}.
	\end{itemize}
	
	\section{Related Works}
    Many attempts on distributed training over multiple machines have recently given rise to research on decentralized machine learning \cite{maDistributedOptimizationArbitrary2017, wangWhenEdgeMeets, reddiAIDEFastCommunication2016}.  However, most of the algorithms in these works are designed for machines having balanced and i.i.d. data, and is connected to high-throughput networks in data centers. With a different scheme, Federated Learning (and related on-device intelligence approaches), which has attracted much attention recently  \cite{konecnyFederatedOptimizationDistributed2016, konecnyFederatedLearningStrategies2016, FederatedLearningCollaborative, mcmahanCommunicationEfficientLearningDeep2017, konecnySemistochasticCoordinateDescent2017, FedProx2020, li2020federated}, exploits the collaboration of mobile devices that can be large in number, slow and/or unstable in Internet connections, and have non-i.i.d. and unbalanced data locally. In the simplest form of a federated learning algorithm (i.e., FedAvg \cite{mcmahanCommunicationEfficientLearningDeep2017}), the authors provide a simple mechanism of averaging the updated learning parameters of the local model using local data at individual UEs. On the other hand, CoCoA+ \cite{konecnySemistochasticCoordinateDescent2017} provides a general framework under a strong theoretical convergence analysis, in which the local learning problem of UEs is transformed into dual problems and can be solved by any arbitrary solvers. Different from CoCoA+ framework, in our recent work \cite{FL_TON2019}, we develop a new federated learning algorithm, namely \FEDL that uses an additional Bregman divergence in the learning objective \cite{banerjeeClusteringBregmanDivergences2005} to encourage the local model being close to the global model.	Most of these existing works aim to provide the learning algorithms that focus on learning performance and providing the theoretical convergence analysis. However, for the practical deployment in mobile edge networks, a resource allocation problem for federated learning service needs to be carefully designed to manage the computation and communication resources. Furthermore, it is important to control the learning parameters by considering energy consumption and learning time convergence. The work of \cite{wangWhenEdgeMeets} introduced this type of problem for the distributed gradient descent analysis, in which the authors propose a control algorithm that determines the best trade-off between local updates and global parameter aggregation to minimize the loss function under a given resource budget. 
    
    In our previous work \cite{tranFederatedLearningWireless2019,FL_TON2019}, we propose a resource allocation problem among UEs and the hyper-learning rate control of learning services in the wireless environment regarding the computation, communication latency, UE energy consumption, and the heterogeneity of UEs. In this paper, we study the extensive design for multiple federated learning services that co-exist at the edge networks with the change in the sharing of bandwidth allocation based on OFDMA instead of transmission power control in our previous work. Also, we consider the additional broadcast time, extra communication overhead, and the time for averaging operation at the edge server. Furthermore, both resource allocation and learning processes can be controlled by a FLO at the MEC server which performs the resource allocation in the centralized or decentralized approaches and is being an aggregator for the global learning update. For the centralized solution approach, the \Opt problem is bi-convex and we adopt the alternative minimization algorithm to provide a solid approach that can help the subproblems of the \Opt problem can be solved by arbitrary convex solvers. When all services from the same owner such as multiple deep learning models can be deployed together and provide better performance mobile vision systems in \cite{DeepyEye2017,Fang2018}, this approach can be sufficient to provide an efficient resource allocation mechanism in which the sharing of service information in the \Opt problem is not an issue. However, the centralized approach will be limited when scaling up to a large number of users and require the sharing of all information among services. To resolve these problems, we develop another flexible decentralized solution approach based on the combination of multi-convex and parallel setting of ADMM. The conventional ADMM algorithm can be applied in convex problem \cite{boyd2011distributed}, and later extended to parallelly solve subproblem with JP-ADMM \cite{jpadmm2017}. Recently, the new analysis for the convergence of the multi-convex ADMM problem \cite{miADMM2019}. To the best of our knowledge, the combination of these two extended algorithms for ADMM hasn't applied in other works as in our decentralized approach. Our decentralized approach can be useful for independent service providers such as multi-tenant FL services that use the common shared resources from the third-party edge provider. The decisions can be made by each service and shared with FLO only without revealing the learning service information (i.e., dataset information, exchange local updates information between UEs and the MEC server, the number of CPU cycles for each UE to execute one sample of data).

	\section{Multi-Service Federated Learning at the Edge} \label{S:MSFL_Model}

\begin{algorithm} [t]
	\caption{Federated Learning Algorithm (\FEDL)\cite{FL_TON2019}}  \label{Alg0}
	\begin{algorithmic}[1]
		\State \tbf{Input:} $w^{0}$, $\theta \in [0,1]$, $\eta>0$. 
		\For {$t =  1 \text{\,to\,}  K_g$}
		\State \multilines{\textbf{Local Training:} Each UE $n$ solves its local learning problem \eqref{E:Computation}	in $K_l$ rounds to achieve $\theta$-approximation solution $w_n^{t}$ satisfying \eqref{E:theta_approximation}.} \label{line:cp}
		\State \multilines{\textbf{Communication:} All UEs transmit $w_n^{t}$  and $\nabla F_n (w_n^{t} )$ to the edge server.}  \label{line:co}
		\State \multilines{\textbf{Aggregation and Feedbacks:} The edge server updates the global model as in \eqref{E:Agg}
			and then fed-backs $w^t$  and  $\nabla F (w^{t})$ to all UEs.} \label{line:bs}
		\EndFor
	\end{algorithmic}
\end{algorithm}

\subsection{Federated Learning Algorithm Design}\label{S:Alogrithm}
In this subsection, we summarize our recent federated learning algorithm design according to \cite{FL_TON2019} as in the detail of Algorithm \ref{Alg0}.
Accordingly, in the typical setting of federated learning for a general supervised learning problem, given a sample data $\{x_i,y_i\}\in\cal{D}$ with input $x_i \in \mathbb{R}^d$, the learning task is required to train the \emph{model parameter} $w\in \mathbb{R}^d$ to predict the correct label $y_i$ by minimizing the loss function $f_i (w)$. The training data at UEs can be the usage information or sensing data from the integrated sensors. Different from the conventional centralized learning, the dataset of the federated learning scheme is distributed over a set of $N$ UEs where each participating UE $n$ collects training data samples and stores a local dataset $\mathcal{D}_{n}$ such that $$\Cal{D} = \underset{n=1..N}{\cup} \Cal{D}_n; \underset{n=1..N}{\cap} \Cal{D}_n =\emptyset. $$  

The local loss function of the learning problem using the local dataset of UE $n$ is defined as
\begin{align}
F_n (w) \defeq \frac{1}{|\Cal{D}_n|} \SumNoLim{i \in \Cal{D}_n}{} f_i ( w). \label{E:Local_Loss}
\end{align}

\begin{assumption} \label{Assumption}
	The local loss function $F_n(\cdot)$ is $L$-smooth and $\beta$-strongly convex, $\forall n$, respectively, as follows, $\forall w, w' \in \mathbb{R}^d$:
	\begin{align*}
	F_n(w)  &\leq F_n(w')  \!+\!  \bigl\langle \nabla F_n (w'), w - w' \bigr\rangle\! + \! \frac{L}{2} \norm{w - w'}^2 \\
	F_n(w)  &\geq F_n(w')  \!+\! \bigl\langle \nabla F_n (w'), w - w' \bigr\rangle  \!+\! \frac{\beta }{2} \norm{w - w'}^2,  
	\end{align*}
\end{assumption}
where $\langle x, x'\rangle$ denotes the inner product of vectors $x$ and $x'$ and $\norm{\cdot}$ is Euclidean norm. These strong convexity, smoothness assumptions are also used in \cite{wangAdaptiveFederatedLearning2019}, and satisfied in popular ML problems such as $l_2$-regularized linear regression model with $f_i (w) = \frac{1}{2} (\innProd{x_i}{w} - y_i)^2 + \frac{\beta}{2} \norm{w}^2, y_i \in \DD{R}$, and $l_2$-regularized logistic regression with  $f_i (w) = \log \bigP{ 1 + \exp ( - y_i \innProd{x_i}{w}) } + \frac{\beta}{2} \norm{w}^2, y_i \in \{-1, 1\}$. Accordingly, we denote $ \rho  \defeq \frac{L}{\beta}$ as the condition number of $F_n(\cdot)$'s Hessian matrix. 

Then, the global loss function of the global learning problem is as follows
\begin{align}
\min_{w \in \mathbb{R}^d} F(w) \defeq \SumNoLim{n=1}{N} \frac{|\Cal{D}_n|}{|\Cal{D}|} F_n (w). \label{E:Global_Loss}
\end{align}

Accordingly, the global learning model can be obtained by solving the global problem \eqref{E:Global_Loss} using an iterative update process at the server and UEs in the federated learning scheme. These updates perform alternatively within a number of \emph{global rounds} (i.e., $K_g$) that consists of four following steps at one global round as
\begin{itemize}
	\item \textit{S1. Local Training}: Every UE needs to train a local model by using the local training data $\mathcal{D}_n$. 
	\item \textit{S2. Upload local model}: UEs transmit the local learning model and global gradient updates to the server. 
	\item \textit{S3. Update global model}: The global model is constructed based on the weight parameters of local models at the server.
	\item \textit{S4. Broadcast global model}: The updated global model and gradient are broadcast to all UEs.
\end{itemize}

\textbf{Local Training at UEs:} According to \cite{FL_TON2019}, instead of solving the local objective in the equation \eqref{E:Local_Loss}, the surrogate problem is solved to attain the local model $w_n^{t}$ for each global round $t$ as follows
\begin{align}			
	\min_{w \in \DD{R}^d} J_n^t (w) \defeq D_{F_n} (w, w^{t-1}) + \eta \,  \hat{F} (w | w^{t-1}),   \label{E:Computation}
\end{align}
where $D_{F_n}$ denotes the Bregman divergence \cite{banerjeeClusteringBregmanDivergences2005} of $F_n(\cdot)$
\begin{align*}			
	&D_{F_n} (w, w^{t-1}) \nonumber \\
	&\quad \defeq  F_{n} (w) -  F_{n} (w^{t-1}) - \innProd{\nabla F_n (w^{t-1} )}{w -w^{t-1}};
\end{align*}
$\hat{F} (w | w^{t-1}) $ denotes the first-order approximation of the global function $F(\cdot)$ at $w^{t-1}$
\begin{align*}			
	\hat{F} (w | w^{t-1}) \defeq  F (w^{t-1}) + \innProd{\nabla F (w^{t-1} )}{w -w^{t-1}};
\end{align*}
 and $\eta > 0$ is the weight that balances between two objectives which is also our controlled learning parameter. The Bregman divergence is the generalized distance between the local model solution $w$ and the latest global model parameter $w^{t-1}$ (e.g., square Euclidean distance) that is widely applied in machine learning applications, statistics, and information geometry \cite{banerjeeClusteringBregmanDivergences2005}. 
 Thus, the local model at UEs can be constructed by minimizing the surrogate objective with the approximated loss function minimization such that its parameters is close to the latest global model parameter $w^{t-1}$. Then the equivalent local learning problem is derived as follows
\begin{align}			
\min_{w \in \DD{R}^d} J_n^t (w) \eqdef F_{n} (w) +  \bigl\langle  \eta \nabla F (w^{t-1}) - \nabla F_n (w^{t-1} ), w \bigr\rangle.   \label{E:Computation0}
\end{align}
 Since it is usually difficult to obtain the optimal solution in the learning problem \eqref{E:Computation0}, UEs is required find a \textit{(possibly weak)} solution $w_n^{t}$ instead. As an analogy from the definition for the relative accuracy in \cite{smithCoCoAGeneralFramework2018, reddiAIDEFastCommunication2016} for the approximation, the local weight parameters at all UEs satisfy 
\begin{align}			
	\norm{\nabla J_{n} (w_n^{t}) } \leq \theta \, \norm{\nabla J_{n} (w^{t-1}) },  \forall n, \label{E:theta_approximation}
\end{align}
where the relative local accuracy $\theta \in (0,1)$ is common to all UEs. This parameter also defines the quality of the approximation solutions when solving the local learning problem \eqref{E:Computation}, in which $\theta = 0$ the optimal solution is obtained, while $\theta = 1$ we have no progress (i.e., $w_n^t = w^{t-1}$).
Since the objectives $J_n^t (w)$ and $F_n(\cdot)$ have the same Hessian matrix, $J_n^t (w)$ is also $\beta$-strongly convex and $L$-smooth. Accordingly, the gradient descent (GD) method is reasonable to solve \eqref{E:Computation0} and requires $K_l$ number of \emph{local iterations} to achieve the accuracy $\theta$-approximation of the solution.
\begin{align}
w^{k+1}_n = w^{k}_n - h_k \nabla J_{n}^t (w^{k}_n ), \label{E:local_gradient}
\end{align}
where $h_k$  is a learning rate. Note that each UE holds a small portion of samples, i.e, $D_n << D$, $\forall n$. In case of large $D_n$, mini-batch SGD can be used to alleviate the computation burden on UEs, but the convergence rate will be different. We assume that the generated convergent sequence $(w^k_n)_{k \geq 0}$ for the local model satisfying a linear convergence rate \cite{nesterovLecturesConvexOptimization2018} as follows
\begin{align}
J_{n}^t (w^k_n)  - J_{n}^{t} (w^*_n) \leq c (1 - \gamma)^k \bigP{J_{n}^t (w_0)  - J_{n}^{t} (w^*_n)}, \label{E:gradient_based}
\end{align}
where $w^*_n$ is the optimal solution of the local problem \eqref{E:Computation0}, and $c$ and $\gamma \in (0,1)$ are constants depending on $\rho$. 
\begin{lemma} \label{Lem:local}
	With Assumption~\ref{Assumption} and the assumed linear convergence rate \eqref{E:gradient_based} with $w_0 = w^{t-1}$, the number of local rounds for solving \eqref{E:Computation} to achieve a $\theta$-approximation condition \eqref{E:theta_approximation}  is
	\begin{align}
	K_l  = \frac{2}{ \gamma} \log \frac{C}{\theta },  \label{E:K_l}
	\end{align}
	where $C \defeq  c \rho $. 
\end{lemma}

\textbf{Global model updates at the server:}
Considering a synchronous federated learning scheme, the global model parameter is then updated by aggregating the local model parameter $w_n^t$ from all UEs as follows
\begin{align}
w^{t} =  \SumN \frac{|\Cal{D}_n|}{|\Cal{D}|}  w_n^{t}. \label{E:Agg}
\end{align}

This updated global model is then broadcast along with  $\nabla F (w^{t})$ to all UEs (line~\ref{line:bs}) to all UEs. The convergence of the global problem \eqref{E:Global_Loss} is achieved by satisfying 
\begin{align}
F(w^{t}) - F(w^*) \leq  \epsilon, \; \forall t \geq  K_g, \label{E:global}
\end{align}
where $\epsilon>0$ is an arbitrarily small constant, $w^*$ is the optimal solution of the problem \eqref{E:Global_Loss}. The algorithm enhances data privacy by exchanging learning information only rather than the local training data. 
The convergence analysis for \FEDL is developed in \cite{FL_TON2019}.
\begin{theorem} \label{Th:1}
	With Assumption~\ref{Assumption}, the convergence of the \FEDL algorithm is achieved with linear rate 
	\begin{align}
	F(w^{t}) - F(w^*) \leq (1-\Theta)^k (F(w^{(0)}) - F(w^*)), \label{E:global_convergence}
	\end{align}
	where $ \Theta \in (0,1)$ is defined as  
	\begin{align}
	\Theta\defeq 	\frac{\eta(2(\theta-1)^2- (\theta+1)\theta(3\eta+2)\rho^2-(\theta+1)\eta\rho^2)}{2\rho\bigP{(1+\theta)^2\eta^2\rho^2 + 1}}. \label{E:Theta} 
	\end{align}
\end{theorem}

\begin{corollary} \label{Co:1}
	The number of global rounds for \FEDL to achieve the convergence satisfying \eqref{E:global} is
	\begin{align}
	K_g  =\frac{1}{\Theta} \log \frac{F(w^0) - F(w^*)}{\epsilon},  \label{E:K_g}
	\end{align}
\end{corollary}

We show the detail proofs of Lemma \ref{Lem:local}, Theorem \ref{Th:1}, and Corollary \ref{Co:1} in the work \cite{FL_TON2019}.

	\textbf{Learning Time Model:}
		According to the convergence analysis of the federated learning algorithm, we obtain the convergence rate and the global rounds depend on the hyper-learning rate $\eta$ and the relative accuracy of the local learning problem $\theta$ as in Corollary \ref{Co:1}. Therefore, the total learning time can be defined in the general form as follows
		\begin{equation}
		\text{TIME}(\eta,\theta) = K_g(\Theta)\times (c+\Cal{T}(\theta)),
		\end{equation}
		where $c$ is the one round of communication time, $\Cal{T}(\theta)$ is the required time to obtain the relative accuracy $\theta$ of the local learning algorithm, and $ K_g(\Theta)$ is the required number of global rounds in the equation \eqref{E:Theta} and the $\Theta$ is defined in the equation \eqref{E:Theta}. In a common setting of many federated learning frameworks \cite{mcmahanCommunicationEfficientLearningDeep2017,FedProx2020}, the number of local iterations is often fixed for each UE, thus, the remaining control parameter is the hyper-learning rate $\eta$ that affects to the number of global rounds. Accordingly, we substitute the constants and get the simplified form as follows
				
		\begin{align}
		&K_g(\Theta) = \frac{A}{\Theta};  \nonumber\\
		&\Theta = \frac{\eta\big(2(\theta-1)^2- 2(\theta+1)\theta\rho^2 -\eta\rho^2(\theta+1) (3\theta+1) \big)}{2\rho\bigP{(1+\theta)^2\eta^2\rho^2 + 1}}, \nonumber\\
		& = 	\frac{C\eta -D\eta^2}{2\rho\bigP{B\eta^2 + 1}}, \label{E:K_Theta}
		\end{align}
		where
		\begin{align}
		A \defeq& \log \frac{F(w^0) - F(w^*)}{\epsilon} >0, \nonumber\\
		B \defeq& (1+\theta)^2\rho^2, \nonumber\\
		C \defeq& 2(\theta-1)^2- 2(\theta+1)\theta\rho^2, \nonumber\\
		D \defeq& \rho^2(\theta+1) (3\theta+1).  \nonumber 
		\end{align}

	\begin{table}[]
		\centering
		
		\caption{\textbf{The summary table of important notations}}
		\label{notation}
		\begin{tabular}{ |p{0.045\textwidth}| p{0.37\textwidth}|}
			\hline
			\textbf{Var}		& \textbf{Definition}              					 	\\ 
			\hline
			\hline
			$s$         		& Index denoting FL service					 	\\
			$n$ 				& Index denoting participating UE		 					 	\\
			$D_{s,n}$			& The size of local dataset of the service $s$ at UE $n$  		 	\\ 
			$c_s$				& The number of CPU cycles required to process 1 bit of data sample of the service $s$ 	\\
			$\Ecomp$			& The energy consumption of service $s$ at UE $n$ to compute one local iteration \\
			$\Ecomm$			& The energy consumption of service $s$ at UE $n$ to transmit the local updates\\
			$\Tcomp$ 			& The local training time for one local iteration of service $s$\\
			$\Tcomm$    		& The transimission time of service $s$\\
			$B^{ul}$			& The total uplink bandwidth \\
			$B^{dl}$			& The downlink bandwidth \\
			$v_s$				& The size of local information updates of the service $s$\\
			$K_{l,s}$			& The number of local iterations of service $s$  		\\
			$K_g(\Theta_s)$		& The number of global rounds using \FEDL \\
			$\mathcal{C}_s$ 	& The total cost of the learning service $s$  \\
			$\eta_s$			& The hyper-learning rate $\eta_s$ using \FEDL \\
			$f_{s,n}$ 			& The allocated CPU-cycle frequency for service $s$ at UE $n$\\
			$w_n$				& The allocated fraction of the uplink bandwidth for UE $n$		 	\\
			\hline
		\end{tabular}
	
	\end{table}
	
	\subsection{Multi-Service Sharing Model}
	In this paper, we consider a multi-service federated learning scheme with one Federated Learning Orchestrator (FLO) at the MEC server and a set $\Cal{N}$ of $N$ UEs as shown in Fig. \ref{F:System_model}. Each participating UE $n$ stores a local data set  $\mathcal{D}_{s,n}$ with size $D_{s,n}$ for each federated learning service $s$. Then, we can define the total data size of a service $s$ by $D_s= \sum_{n=1}^{N} D_{s,n}$. The CPU resource of each UE is consumed to perform the local learning problem in \eqref{E:Computation0} for each service $s$ by using the local data. Therefore, it is crucial to share the CPU resource of each UE amongst the local learning problems of multiple services efficiently. After the local training, all UEs upload their updated local model parameters to the MEC server by using the wireless medium. Hence, it is also important to efficiently share the communication resource (i.e., bandwidth) among the UEs. At the MEC server, FLO is deployed to manage computation (i.e., CPU) and communication resources sharing among learning services and UEs. In addition to resource allocation, FLO also controls the hyper-learning rate of learning services.
	
	\subsubsection{Local Computation Model}
	We denote the required number of CPU cycles for each UE to execute one sample of data belong to service $s$ by $c_s$, which can be measured offline \cite{miettinenEnergyEfficiencyMobile2010}. The required CPU cycles are directly proportional to the number of samples in the local dataset. Since all samples $\{x_i, y_i\}_{i \in \Cal{D}_{s,n}}$ have the same size (i.e., number of bits), the number of CPU cycles required for UE $n$ to run one local iteration of learning service $s$ is $c_s  D_{s,n}$. The allocated CPU-cycle frequency for the service $s$ is denoted by $f_{s,n}$. Then the energy consumption of UE $n$ to compute one local iteration for learning service $s$ can be expressed as follows  \cite{burdProcessorDesignPortable1996}
	\begin{align}
	\Ecomp = \SumNoLim{i=1}{c_s D_{s,n}} \frac{\beta_n }{2} f_{s,n}^2 = \frac{\beta_n }{2} c_{s} D_{s,n} f_{s,n}^2, \label{E:Ecomp}
	\end{align}
	where $\beta_n/2$ is the effective capacitance coefficient of UE $n$'s computing chipset. In addition, the computation time per local iteration of the UE $n$ for a service $s$ is $ \frac{c_{s} D_{s,n}}{f_{s,n}}.$ Using a synchronous federated learning scheme, local training time for one local iteration of the learning service $s$ is the same with the computation time of the slowest UE as
	
	\begin{equation}
	\Tcomp = \max_{n \in \Cal{N}} \frac{c_{s} D_{s,n}}{f_{s,n}} + \tau^{m}_{s,n},\label{E:Tcomp}
	\end{equation}
	where $\tau^{m}_{s,n}$ is the extra overhead to access memory. We denote the vector of $f_{s,n}$ by $f_s\in \DD{R}^n.$ 
	
	
	\subsubsection{Communication Model}
	After processing the local learning problem, UEs need to exchange the local information to FLO on a shared wireless medium via a multi-access protocol (i.e., OFDMA). Therefore, the achievable transmission rate (\textit{bps}) of UE $n$ on the given the allocated fraction $w_n$ of the total uplink bandwidth $B^{ul}$ is defined as follows:
	\begin{align}
	r^{ul}_n (w_n) = w_n B^{ul} \log_2 \bigP{ 1 + \frac{h_n p_n}{N_0}}, \label{E:rate1}
	\end{align}  
	where $N_0$ is the background noise, $p_n$ is the transmission power, and $h_n$ is the channel gain of the UE $n$. Since the dimensions of local models and global gradient updates in line 4 of the Alg. \ref{Alg0} are fixed for all UEs, the data size (in bits) of local information updates does not change and is denoted by $v_s$ for each learning service $s$.  
	Thus, uplink transmission time of each UE $n$ for a service $s$ is 
	\begin{align}
	\tau^{ul}_{s,n}(w_n) = \frac{v_{s}}{r^{ul}_n(w_n)}. \label{E:rate2}
	\end{align}
	
	In addition, the downlink broadcast delay should be taken into account to transmit the global changes to all users using the downlink bandwidth $B^{dl}$ as follows
	\begin{align}
	r^{dl}_n = B^{dl} \log_2 \BigP{ 1 + \frac{h_n P}{N_0}}. \label{E:rate3}
	\end{align}
	Since the global information and the local information has the same size $v_s$, then, the downlink transmission time for the updated global information is $ \tau^{dl}_{s,n} = \frac{v_{s}}{r^{dl}_n}.$ Thus, the downlink of the service $s$ is $\tau^{dl}_{s} = \max_{n\in \Cal{N}} \tau^{dl}_{s,n}$.
	
	Accordingly, the communication time of a learning service $s$ consisting of the uplink transmission time and downlink transmission time is defined as

	\begin{equation}
	\Tcomm = \max_{n\in \Cal{N}} \tau^{ul}_{s,n}(w_n) + \tau^{dl}_{s} + \tau^{ex}_{s,n}, \label{E:Tcomm}
	\end{equation}
	where $\tau^{ex}_{s,n}$ is the extra communications overhead during the transmission (e.g., establishing TCP connection) and assumed to be random constants for each FL service communication.
	
	Furthermore, the energy consumption for the uplink communication of UE $n$ for the service $s$ is defined as
	$$\Ecomm = p_n\tau^{ul}_{s,n}(w_n).$$
	
	\subsubsection{Global Round Model}
	As aforementioned in subsection II.A, we have the number of global rounds and the number of local iterations are $K_g(\Theta)$ and $K_{l,s}$, respectively. 
	Then, the running time of one global round of the learning service $s$ includes local learning time and transmission time which is defined as follows
	
		\begin{align}
		\Titer(\Tcomp, \Tcomm) \defeq  \Tcomm + T^{avg}_{s} +  K_{l,s} \Tcomp,  
		\end{align}
	where  $T^{avg}_{s}$ is the computation time of the averaging operation. Since the simple averaging operation can be performed very quickly with strong computation capabilities of the edge server and the size of local updates information (i.e., local learning model, global gradient updates) are the same for every global round, $T^{avg}_{s}$ is assumed to be small constant for each service $s$.
	
	Furthermore, in one global round of each service $s$, the total energy consumption of all UEs for learning and uplink transmission is expressed as follows
	\begin{align}
	\Eiter (f_s, w) \defeq \SumNoLim{n=1}{N} \Ecomm + K_{l,s} \Ecomp.
	\end{align}
	
	Finally, the total cost of a learning service $s$ is defined as
	\begin{align}
	\mathcal{C}_s \defeq K_g(\Theta_s) \Big(\Eiter (f_s, w) + \kappa_s \Titer (\Tcomp, \Tcomm)\Big), \nonumber
	\end{align}
	where $\kappa_s$ is the trade-off between running time and the energy consumption of UEs that needs to be minimized. 
	%
	
	\subsection{Problem formulation}
	Since the learning services jointly occupy the shared CPU resource in each UE, and the shared uplink bandwidth resource to upload the weight parameters of local models. Thus, FLO takes a role to manage these shared resources and controls the hyper-learning rate of learning services. To minimize the running time cost and the energy consumption of UEs, we propose the multi-service Federated Learning optimization problem for FLO, \FL, as follows
	\begin{align}	
	& \underset{{f,w,\eta}}{\text{min.}} && \sum_{s\in \Cal{S}}  K_g(\Theta_s) \Big(\Eiter (f_s, w) + \kappa_s \Titer (\Tcomp, \Tcomm)\Big)  \nonumber \\
	& \text{s.t.}
	&& \sum_{s\in \cal{S}} f_{s,n} = f_{n}^{tot}, \forall n\in \Cal{N}, (\textbf{\textit{Shared CPU}}) \label{E:constr_sharedCPU} \\
	&&& \sum_{n\in \Cal{N}} w_{n} = 1, \,  (\textbf{\textit{Shared Bandwidth}})\label{E:constr_sharedBW}\\
	&&& f_{s,n} \geq f_{s,min}, \forall s\in \Cal{S}, \forall n \in \Cal{N}, \label{E:constr_cpu}\\
	&&& w_{n}\geq w_{min}, \forall n\in \Cal{N}, \label{E:constr_BW}\\ 
	&&&   0 < \Theta_s <  1; \eta_s>0, \forall s\in \Cal{S}, (\textbf{\textit{Learning parameters}})\label{E:constr_theta}\\ 
	&&& \Tcomp \geq \frac{c_{s} D_{s,n}}{f_{s,n}} + \tau^{m}_{s,n}, \forall s\in \Cal{S}, \forall n\in \Cal{N},\label{E:constr_Tcomp}\\ 
	&&& \Tcomm \geq \tau^{ul}_{s,n}(w_n) + \tau^{dl}_{s} + \tau^{ex}_{s,n}, \forall s\in \Cal{S}, \forall n\in \Cal{N}, \label{E:constr_Tcomm}
	\end{align}	
	where $f_n^{tot}$  is the total CPU frequency of UE $n$. The main decision variables include the allocated CPU frequency (i.e., ${f}:=\{f_{s,n}\}$) for each service $s$ at UE $n$, the allocated fraction  (i.e., ${w}:=\{w_n\}$ ) of total uplink bandwidth for each UE $n$, and the relative accuracy (i.e., ${\theta}:=\{\theta_s\}$) of the local learning problem at UEs. According to the constraints in \eqref{E:constr_cpu}, \eqref{E:constr_BW}, all of the learning services and UEs are required to be allocated at least the minimum amount of CPU frequency and bandwidth to train and upload the learning parameters of local model. Besides, the auxiliary variables $\Tcomp$ is the computational time of one local iteration, and $\Tcomm$ is the uplink transmission time which depends on $f$ and $w$ in the constraint \eqref{E:constr_Tcomp}, \eqref{E:constr_Tcomm}, respectively. The constraint \eqref{E:constr_sharedCPU} indicates the shared CPU resource among learning services and the constraint \eqref{E:constr_sharedBW} defines the shared uplink bandwidth for each UE. Lastly, the constraint \eqref{E:constr_sharedBW} provides the feasible ranges of the learning parameters in the \FEDL algorithm for each learning service.  

	\section{Solutions to \Opt} \label{S:Sols}

	\subsection{Centralized Approach}
	Even though the \FL problem is non-convex, we later show the specific form of this problem is bi-convex \cite{Xu2013}. The convexity proofs of subproblems are shown in the appendices.
	For solving this type of problem, we adopt the popular technique, Block Coordinate Descent (BCD) algorithm \cite{Xu2013}. The block coordinate descent method cyclically solves the optimization problem for each block of variables while fixing the remaining blocks at their last updated values by following the Gauss-Seidel update scheme. 
	In the \FL problem, there are two blocks of variables such as block of $\theta$ and block of $f,w$. The detail of the centralized algorithm is shown in Algorithm \ref{Cen_alg}.

	Given the fixed values $\hat{f},\hat{w}$ for the resource allocation variable block $f,w$ and the corresponding computation, communication time $\hat{\Tcomp}, \hat{\Tcomm}$, the total cost for each learning service $s$ (i.e.,  $\hat{\cal{C}}_s \defeq \Eiter (\hat{f}_s, \hat{w}) + \kappa_s \Titer (\hat{\Tcomp}, \hat{\Tcomm})$), we have the learning parameter decision problem as follows
	
	\SubLearningA\!: \textbf{Learning Parameter Decision Problem} 
	%
	\begin{align}
	&\underset{{\eta}}{\text{min.}} &&  \sum_{s\in \Cal{S}}  K_g(\Theta_s) \hat{\cal{C}}_s  \nonumber \\
	& \text{s.t.} &&  0 <  \Theta_s <  1, \forall s\in \Cal{S}, \nonumber \\
	& && \eta_s > 0, \forall s\in \Cal{S}, \nonumber
	\end{align}	
	where ${\eta}:=\{\eta_{s}\}$. In $K_g(\Theta_s)$, $\Theta_s$ are the functions of the hyper-learning rate $\eta_s$ and it is defined in the equation \eqref{E:K_Theta}.
	However, the hyper-learning rate decision subproblem \SubLearningA can be decentralized for each learning service $s$ without any coupling among services in the constraints. Thus, each service $s$ can make the decision independently by solving the following decentralized subproblem.\\
	
	\SubLearningB\!: \textbf{Decentralized Learning Parameter Decision} 
	\begin{align}
	&\underset{{\eta_s}}{\text{min.}} &&  K_g(\Theta_s) \hat{\cal{C}}_s \nonumber \\
	& \text{s.t.} &&  0 <  \Theta_s <  1 ,\label{E:Theta_constr} \\
	& && \eta_s > 0. \label{E:eta_constr}
	\end{align}	

	\begin{lemma} \label{L:3}
		
		There exists a unique solution $\theta^*$ of the convex problem \SubLearningB \, satisfying the following equation:
		$$\eta_s^* = \frac{-D + \sqrt{D^2 +BC^2 }}{BC},$$ 
		where the relative accuracy of the local learning problem sufficient small and closed to $0$.
	\end{lemma}

	\begin{algorithm}[t]
		\caption{ Centralized algorithm for \FL} 
		\label{Cen_alg} 
		\begin{algorithmic}[1]
			\State FLO updates the information of learning service requirement, UE resources;
			\State Compute $\eta^*$ from \textbf{\SubLearningA problem} using Lemma \ref{L:3};
			\State Compute $f^*$ from \textbf{\SubCPUA problem} given $\eta^*$;
			\State Compute $w^*$ from \textbf{\SubBWA problem} given $\eta^*$;
		\end{algorithmic}
	\end{algorithm}

	\begin{figure}[t]
		\centering
		\includegraphics[width=0.79\linewidth]{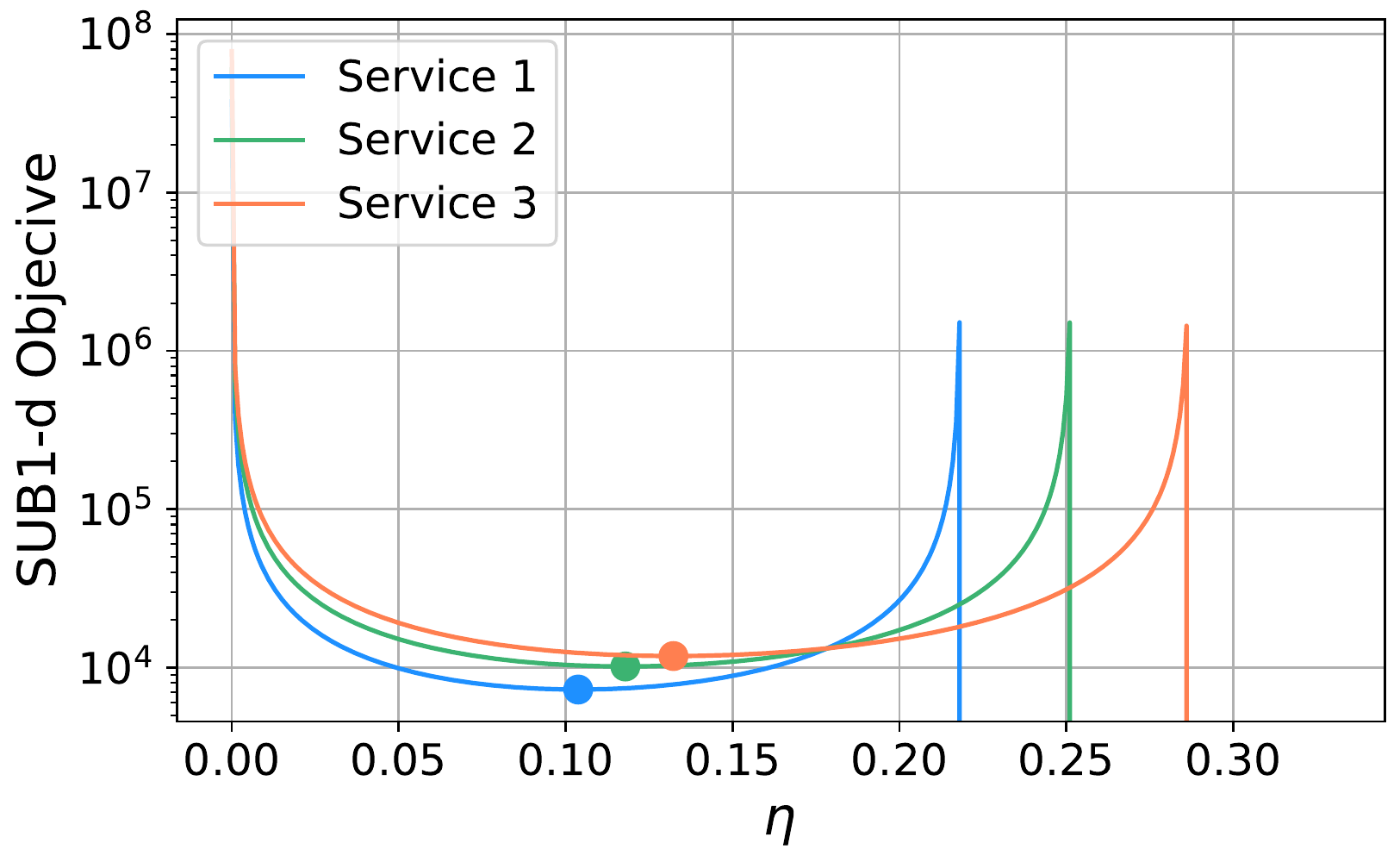}
		\caption{\SubLearningB convexity for three learning services using the similar settings in the next section.}
		\label{F:Sub1_obj}
	\end{figure}
	
		Accordingly, we provide the proof for Lemma \ref{L:3} in the appendix section. In addition, Fig. \ref{F:Sub1_obj} illustrates the convexity of the \SubLearningB subproblem for three learning services. The optimal solutions are obtained by using Lemma \ref{L:3} and marked as the circles which are also the lowest values of the objective curves. As an observation, both high and low values of the hyper-learning rate $\eta$ cause a higher number of global rounds $K_g$ and higher total cost for each learning service.
	
	According to Lemma \ref{L:3}, the optimal hyper-learning rate solutions do not depend on the total cost for each learning service $\hat{\cal{C}}_s$. Thus, the optimal solution of this problem is independent to the other decisions. Then, given the optimal learning parameter $\eta^*$ and the corresponding $\Theta^*_s$, the problem can be decomposed into two independent sub-problems for CPU frequency allocation and bandwidth allocation as follows\\
	
	\SubCPUA\!: \textbf{CPU Allocation Problem} 
	\begin{align}	
	& \underset{{f,T^{cmp}}}{\text{min.}} &&  \sum_{s\in \Cal{S}}  K_{l,s}K_g(\Theta^*_s) \Big( \sum_{n\in \Cal{N}}\frac{\beta_n }{2} c_{s} D_{s,n} f_{s,n}^2 + \kappa_s T^{cmp}_s \Big)  \nonumber\\
	& \text{s.t.}
	&& \Tcomp \geq \frac{c_{s} D_{s,n}}{f_{s,n}} + \tau^{m}_{s,n}, \forall s\in \Cal{S}, \forall n\in \Cal{N},\nonumber\\
	&&& \sum_{s\in \cal{S}} f_{s,n} = f_{n}^{tot}, \forall n\in \Cal{N}, (\textbf{\textit{Shared CPU}}) \nonumber \\
	&&& f_{s,n} \geq f_{s,min}, \forall s\in \Cal{S}, \forall n \in \Cal{N}, \nonumber
	\end{align}
	where ${T^{cmp}}:=\{\Tcomp\}$. \\
	This problem decides a number of CPU frequency for each learning service at UEs.\\
	
	\SubBWA\!: \textbf{Bandwidth Allocation Problem} 
	\begin{align}	
	& \underset{{w,T^{com}}}{\text{min.}} &&  \sum_{s\in \Cal{S}}  K_g(\Theta^*_s) \Big( \sum_{n\in \Cal{N}}p_n\tau^{ul}_{s,n}(w_n) + \kappa_s T^{com}_s \Big) \nonumber\\
	& \text{s.t.}
	&& \Tcomm \geq \tau^{ul}_{s,n}(w_n) + \tau^{dl}_{s} + \tau^{ex}_{s,n}, \forall s\in \Cal{S}, \forall n\in \Cal{N}, \nonumber\\
	&&& \sum_{n\in \Cal{N}} w_{n} = 1, \,  (\textbf{\textit{Shared Bandwidth}}) \nonumber\\
	&&& w_{n}\geq w_{min}, \forall n\in \Cal{N},\nonumber 
	\end{align}
	where ${T^{com}}:=\{\Tcomm\}$. 	
		
	\begin{figure}[t]
		\centering
		\includegraphics[width=\linewidth]{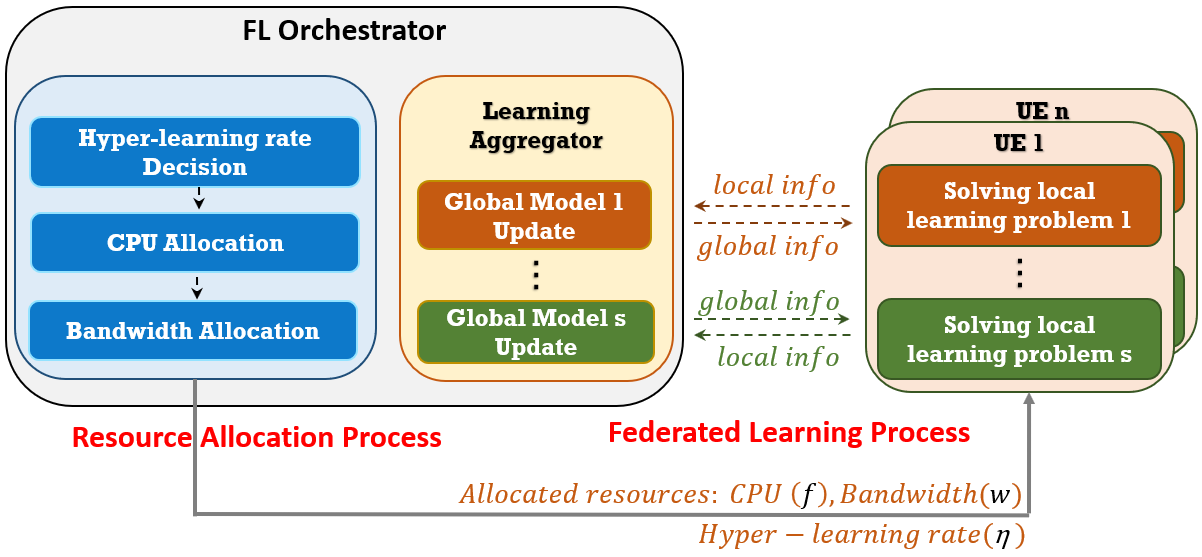}
		\caption{Centralized Deployment.}
		\label{F:Centralized_Deployment}
	\end{figure}	

	\begin{figure*}[t]
		\centering
		\includegraphics[width=0.9\linewidth]{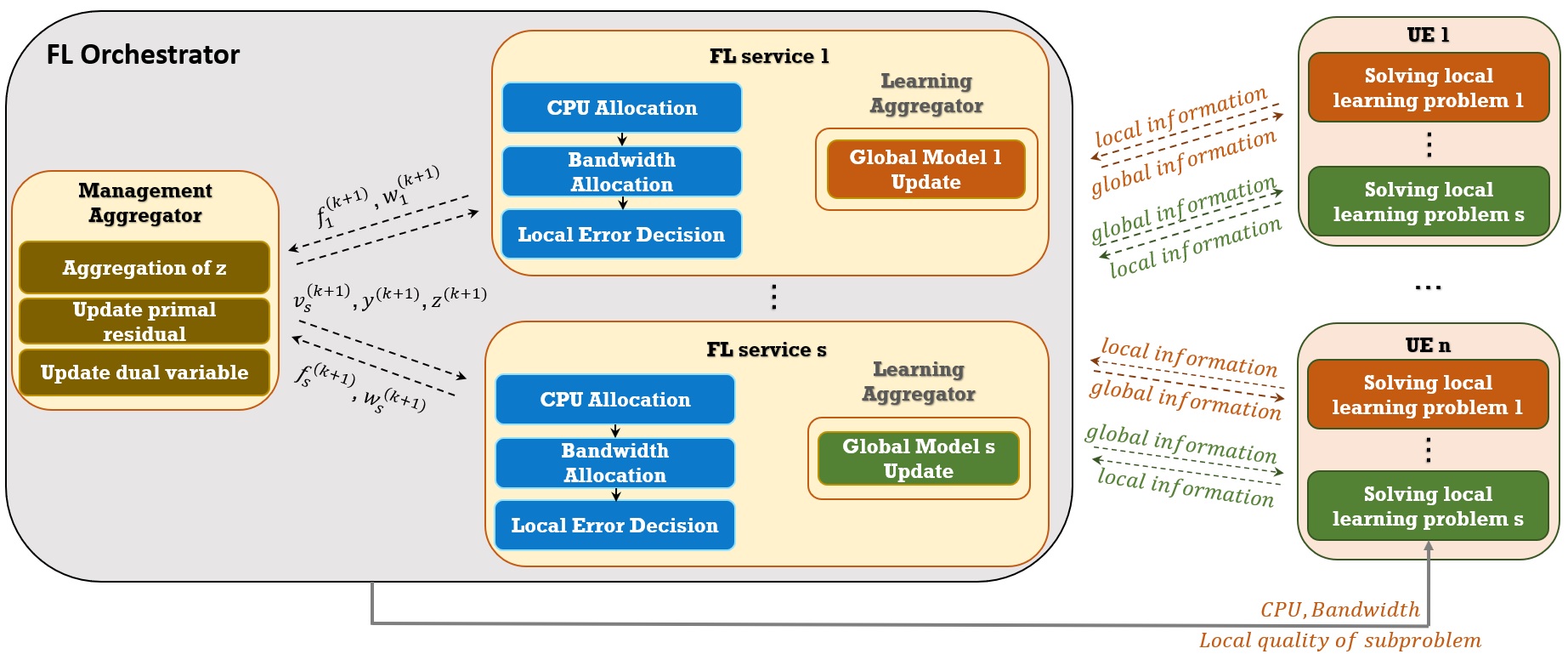}
		\caption{Decentralized Deployment.}
		\label{F:Decentralized_Deployment}
	\end{figure*}
	
	Using block coordinate descent, FLO solves alternatively three convex subproblems of the hyper-learning rate decision, CPU allocation, bandwidth allocation. Discussion as mentioned above, the optimal hyper-learning rate decision can be obtained independently, then only one iteration is required in block coordinate descent style algorithm. These convex problems that can be easily solved by using a off-the-shelf convex solver (i.e., IpOpt solver \cite{wachter2006implementation}). After solving these problems, the resource allocation solutions and the decision of the hyper-learning rate are sent to the UEs to start the learning process. The whole operation process is illustrated in Fig. \ref{F:Centralized_Deployment}.

	\subsection{Decentralized Approach}
	In addition to the centralized algorithm, we develop a decentralized algorithm, which leverages the parallelism structure for subproblems update of Jacobi-Proximal ADMM\cite{jpadmm2017} into the multi-convex ADMM framework\cite{miADMM2019}, namely JP-miADMM. Since the original form of multi-convex ADMM using the conventional Gauss-Seidel scheme does not allow solving the CPU allocation subproblem independently, the integrated Jacobi-Proximal ADMM form provides the parallelism structure for this subproblem. The JP-miADMM algorithm consists of two procedures, such as primal update which can be independently solved by each service $s$ and dual update takes the role of a coordinator from the solutions of learning services. Note that in the following primal subproblems of JP-miADMM, the objectives comprise the addictive norm-2 terms which are the augmented term that originally is introduced in ADMM and proximal term in Jacobi-Proximal ADMM. These two updates are performed iteratively until the convergence condition is obtained. In this algorithm, we introduce the dual variables $v, y$, and the auxiliary variable $z$ that are used in the next subproblems. The detail of the decentralized algorithm is shown in Algorithm \ref{Dec_alg} by alternatively updating the primal variables, primal residual and dual variables until the convergence conditions in line 14 are obtained. Specifically, the first condition is the condition of CPU allocation based on the Frobenius norm of allocation matrix $f$ while the second one is based on the vector norm of bandwidth allocation solution $w$. 
	
	Since the optimal hyper-learning rate decision (i.e., $\eta^*$) is obtained independently according to the closed-form in Lemma \ref{L:3} for each learning service, the JP-miADMM algorithm consists of an iterative process on the shared CPU and bandwidth allocation as follows  

	\subsubsection{Primal Update} In the primal update, each service $s$ solves iteratively its CPU allocation, bandwidth allocation, and hyper-learning rate decision subproblems.\\
	
	\SubCPUB\!: \textbf{Decentralized CPU Allocation} (finding $f_s^{(k+1)}$)
	\begin{align}	
	& \underset{T^{cmp}_s, f_s}{\text{min.}} &&  K_{l,s}K_g(\Theta^*_s) \Big( \sum_{n\in \Cal{N}} \frac{\beta_n }{2} c_{s} D_{s,n} f_{s,n}^2 + \kappa T^{cmp}_s \Big) +{y}^{(k)T}f_{s}  \nonumber\\
	& && + \frac{\rho_1}{2} \big\| \sum_{j \neq s, j \in \Cal{S} }  f_{j}^{(k)} + f_s - f^{tot} \big\|^2 + \frac{\nu}{2} \| f_s -  {f}_s^{(k)}\|^2  \nonumber\\
	& \text{s.t.}
	&& \Tcomp \geq \frac{c_{s} D_{s,n}}{f_{s,n}} + \tau^{m}_{s,n}, \forall n\in \Cal{N},\nonumber\\
	&&& f_{s,n} \geq f_{s,min}, \forall n \in \Cal{N}, \nonumber
	\end{align}
	where $\Theta^*_s$ is the function of $\eta^*$ and the decision variable $f_s$ is the vector of $\{f_{s,n}\}$. Under the mild conditions, i.e., the splittable objective functions are closed proper convex and the existence of a saddle point satisfying KKT condition, 
	the sufficient condition of JP-ADMM for the global convergence to the saddle point according to Theorem 2.1 in \cite{jpadmm2017} can be guaranteed by choosing parameters such that
	$$\nu > \rho_1\Big(\frac{|\mathcal{S}|}{2-\alpha} - 1\Big), \text{ and } 0< \alpha < 2,$$

	\begin{algorithm}[t]
	\caption{ Decentralized algorithm for \FL} 
	\label{Dec_alg} 
	\begin{algorithmic}[1]
		\State FLO updates the information of learning service requirement, UE resources;
		\State Each learning service $s$ computes $\eta^*_s$ from Lemma \ref{L:3};
		\State Initialize $k=1,f^{(1)}, w^{(1)}$;
		\Repeat
		\State \textbf{Primal update}:
		\For{learning service $s \in \mathcal{S} $}	
		\State Compute $f_s^{(k+1)}$ from \textbf{\SubCPUB problem} given $\eta^*_s,f_s^{(k)}$;
		\State Compute $w_s^{(k+1)}$ from \textbf{\SubBWB problem} given $\eta^*_s,z^{(k)}$;
		\EndFor
		\State \textbf{Dual update}:
		\State Update the global consensus bandwidth allocation variable $z^{(k+1)}$ in the equation \eqref{E:z_update};
		\State Update primal residual in the equation \eqref{E:pri_res1}, \eqref{E:pri_res2};
		\State Update dual variable in the equation \eqref{E:dual1}, \eqref{E:dual2};
		\State $k=k+1$;
		\Until{$\|f^{(k+1)} - f^{(k)}\|_F \leq \epsilon_1$, $\|z^{(k+1)} - z^{(k)}\| \leq \epsilon_2$}.
	\end{algorithmic}
\end{algorithm}

	In the bandwidth allocation subproblem, there is a coupling among services due to all services have the same allocated bandwidth allocation $w_n$ at UE $n$. Thus, we transform the \SubBWA subproblem into the following equivalent problem by introducing the auxiliary consensus variable (i.e., ${z} :=\{z_n\}$) and the consensus constraint \eqref{E:constr_consensus}. This transformation is commonly used to handle global consensus variables in ADMM framework \cite{boyd2011distributed}.
	\begin{align}	
	& \underset{T^{com}_s, {w,z}}{\text{min.}} &&  \sum_{s\in \Cal{S}}  K_g(\Theta^*_s)  \Big( \sum_{n\theta}p_n\tau^{ul}_{s,n}(w_{s,n}) + \kappa T^{com}_s \Big) \nonumber\\
	& \text{s.t.}
	&& \Tcomm \geq \tau^{ul}_{s,n}(w_{s,n}) + \tau^{dl}_{s} + \tau^{ex}_{s,n},\forall s\in \Cal{S},  \forall n\in \Cal{N}, \nonumber\\
	&&& \sum_{n\in \Cal{N}} w_{s,n} = 1, \, \forall s\in \Cal{S},  (\textbf{\textit{Shared Bandwidth}}) \nonumber\\
	&&& w_{s,n} \geq w_{min}, \forall s\in \Cal{S}, \forall n\in \Cal{N},\nonumber \\
	&&& w_{s,n} = z_n, \forall s\in \Cal{S},\forall n\in \Cal{N} \label{E:constr_consensus}.  
	\end{align}	
	
	Accordingly, each service $s$ decides the allocated bandwidth $w_s^{(k+1)}$ by solving individually its subproblem as follows\\

	\SubBWB\!: \textbf{Decentralized Bandwidth Allocation} 
	\begin{align}	
	& \underset{T^{com}_s,{w_{s}}}{\text{min.}} &&  K_g(\Theta^*_s)  \Big( \sum_{n\in \Cal{N}}p_n\tau^{ul}_{s,n}(w_{s,n})+ \kappa T^{com}_s \Big)  \nonumber\\
	& && +\nu^{(k)T}({w_{s}} - {z}^{(k)})  + \frac{\rho_2}{2} \big\| {w_{s}} - {z}^{(k)} \big\|_2^2  \nonumber\\
	& \text{s.t.}
	&& \Tcomm \geq \tau^{ul}_{s,n}(w_{s,n}) + \tau^{dl}_{s} + \tau^{ex}_{s,n},  \forall n\in \Cal{N}, \nonumber\\
	&&& \sum_{n\in \Cal{N}} w_{s,n} = 1, \,  (\textbf{\textit{Shared Bandwidth}}) \nonumber\\
	&&& w_{s,n} \geq w_{min}, \forall n\in \Cal{N},\nonumber
	\end{align}	
	where ${w_{s}}:=\{w_{s,n}\}$.
	
	The optimal solution of these convex problems can be easily obtained by using a convex solver (i.e., IpOpt solver \cite{wachter2006implementation}).
	
	\subsubsection{Dual Update} 
	After independently updating the resource allocation, hyper-learning rate decision for each learning service, the dual update is performed to coordinate these solutions and update the dual variables for the next iteration.
	We first update the global consensus variable $z_n$ of the allocated bandwidth for each UE as follows
	\begin{align}
	z_n^{(k+1)} = \frac{1}{|\Cal{S}|} \sum_{ s\in \Cal{S}} \big(w_{s,n}^{(k+1)} + (1/\rho_2) \nu_s^{(k+1)}\big), \forall n \in \Cal{N} \label{E:z_update}.
	\end{align}

	\textit{\\Update primal residual:}
	\begin{align}
	r_1^{(k+1)} &=  \sum_{ s\in \Cal{S} }  f_{s}^{(k+1)} - f^{tot},  \label{E:pri_res1}\\
	r_{2,s}^{(k+1)} &=  w_{s}^{(k+1)} - z^{(k+1)}, 	\label{E:pri_res2}
	\end{align}	
	where $r_1, r_2$ is the vector of $N$ devices.
	
	\textit{Update dual variable:}
	\begin{align}
	y^{(k+1)} &=  y^{(k)} + \rho_1 r_1^{(k+1)}, \label{E:dual1} \\
	\nu_s^{(k+1)} &=  \nu_s^{(k)} + \rho_2 r_{2,s}^{(k+1)}, \label{E:dual2} 
	\end{align}	
	
	where $y:= \{y_n\}, \nu_s:=\{\nu_{s,n}\}$.

	Using JP-miADMM, FLO needs Management Aggregator and a particular module for each learning service. First, each FL learning service module performs CPU allocation, bandwidth allocation, and hyper-learning rate decision then sends the CPU and bandwidth allocation decision to Management Aggregator and then running the aggregation process for variable $z$, primal residual, and dual variables. This process iteratively performs until the convergence condition is achieved. Then, the resource allocation solutions and the decision of the hyper-learning rate are sent to the UEs that participate in the learning process. The whole process of the algorithm deployment is illustrated in Fig. \ref{F:Decentralized_Deployment}. In the MEC server, each service can run its own virtual instance to manage the resource allocation and learning aggregator. Although the decentralized approach requires many iterations to convergence, it provides a more flexible and scalable approach for the resource allocation without revealing the learning service information (i.e., dataset information, the learning weight parameters exchange between UEs and the MEC server, the number of CPU cycles for each UE to execute one sample of data). The decisions can be made by each service and shared with FLO only.
	
\begin{figure}[t]
	\includegraphics[width=0.95\linewidth]{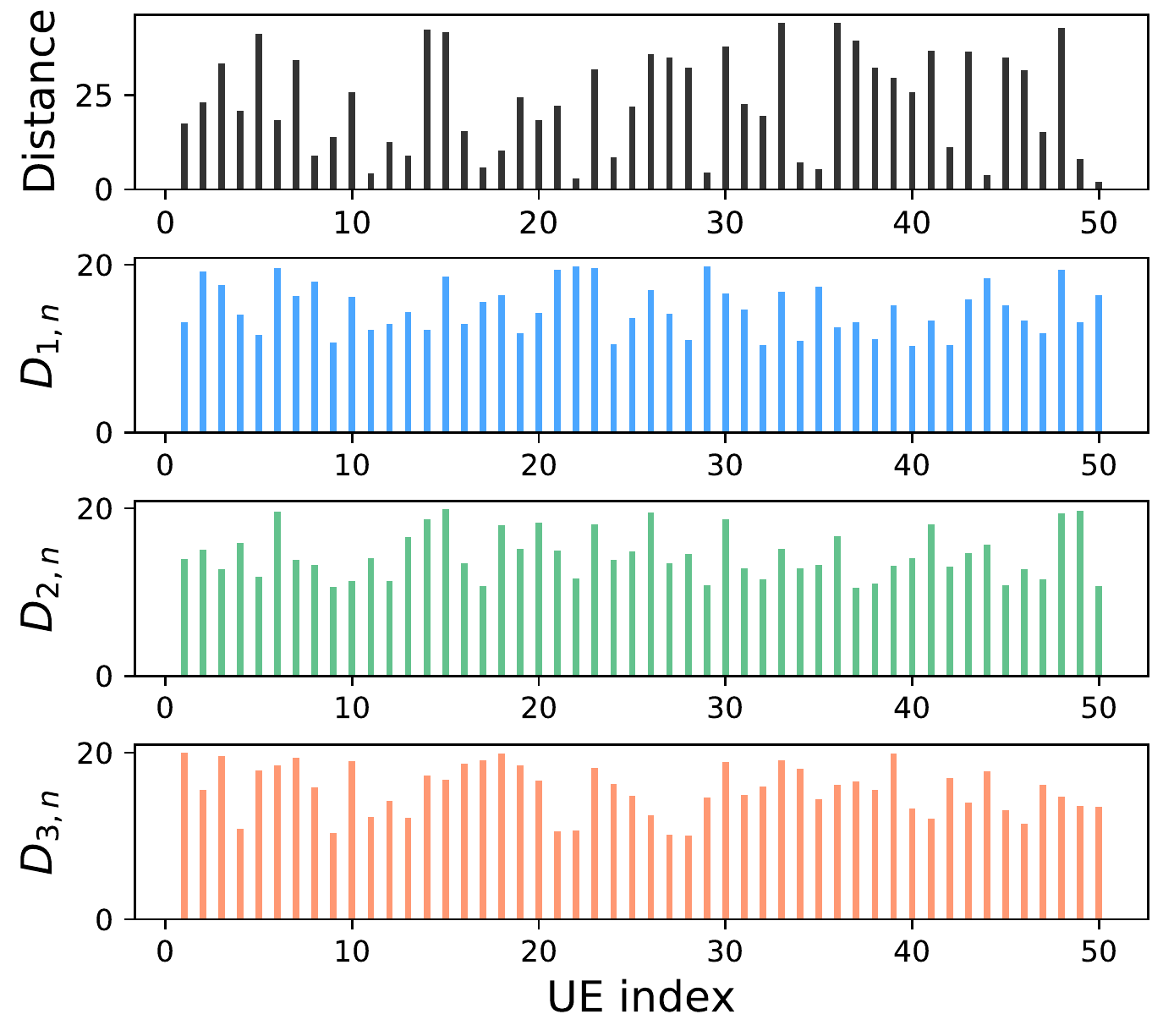}
	\caption{Generated distance (in m) and training size (in MB) at each UE.}
	\label{F:Distance_DataSize}
\end{figure}

\begin{figure*}[t]
	\centering
	\begin{subfigure}{0.325\linewidth}
		\includegraphics[width=\linewidth]{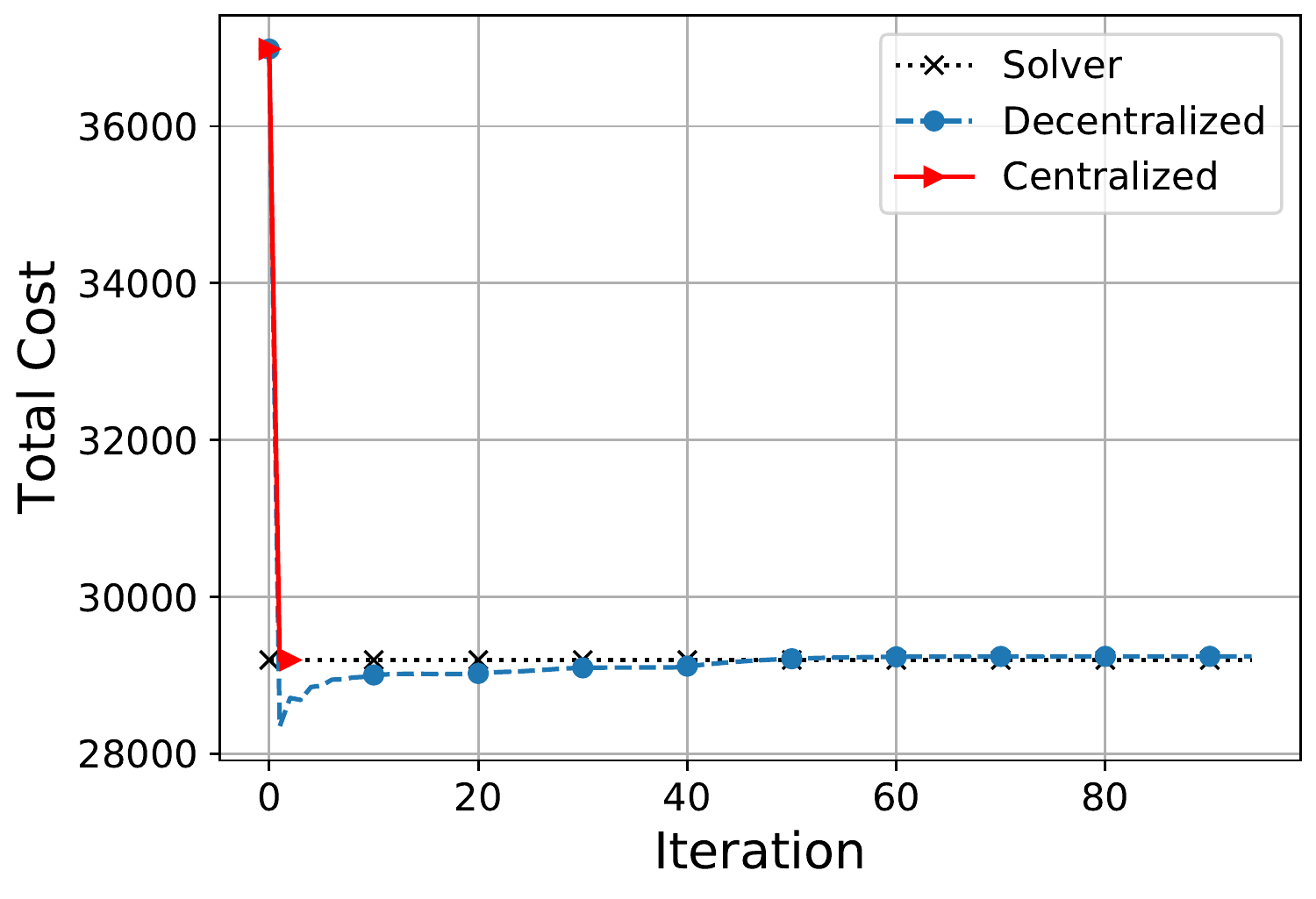}
		\caption{Objective.}
		\label{F:Objective}
	\end{subfigure}
	\begin{subfigure}{0.325\linewidth}
		\includegraphics[width=\linewidth]{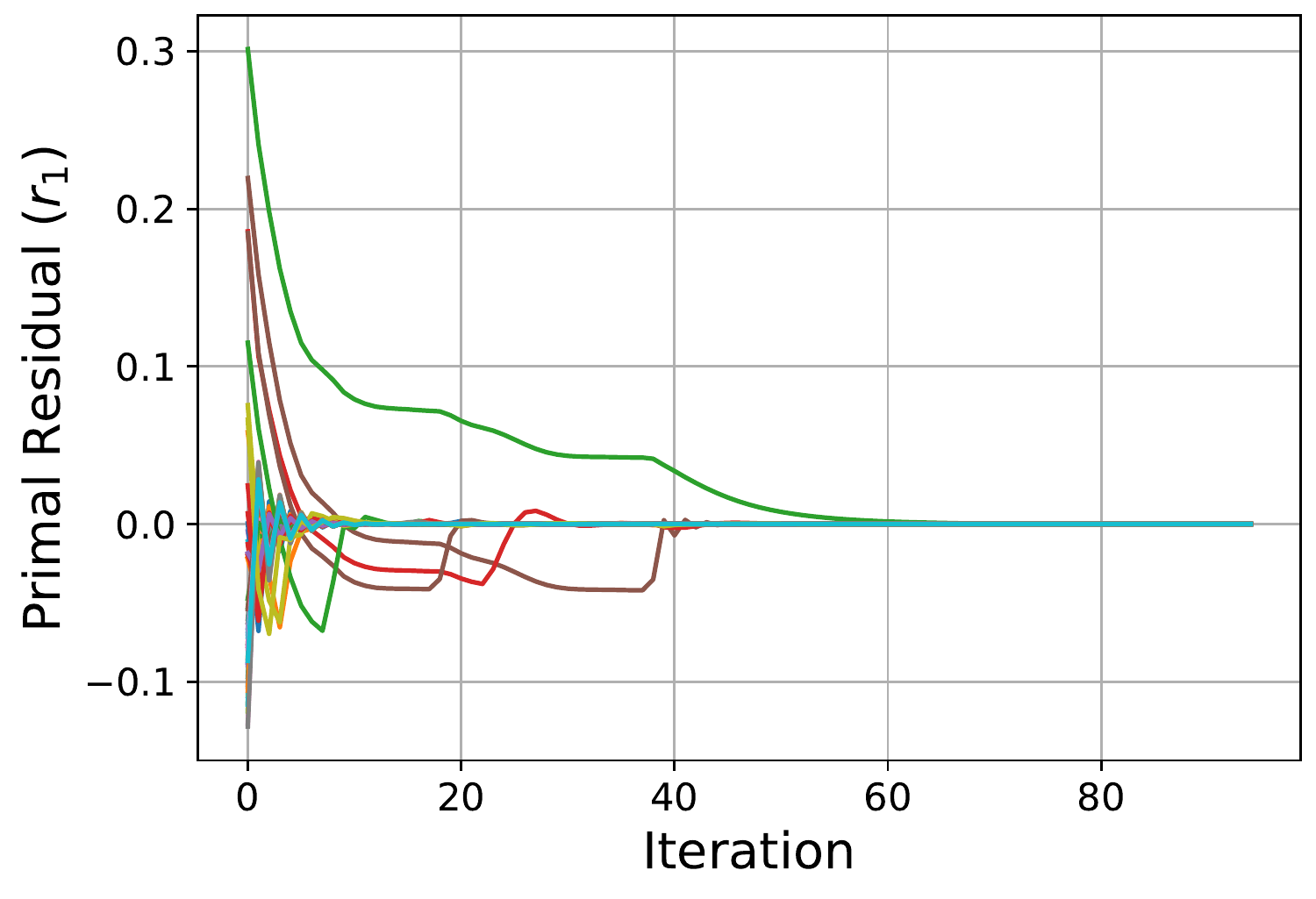}
		\caption{Primal residual $r_1$.}
		\label{F:Convergence_Residual1}
	\end{subfigure}
	\begin{subfigure}{0.325\linewidth}
		\includegraphics[width=\linewidth]{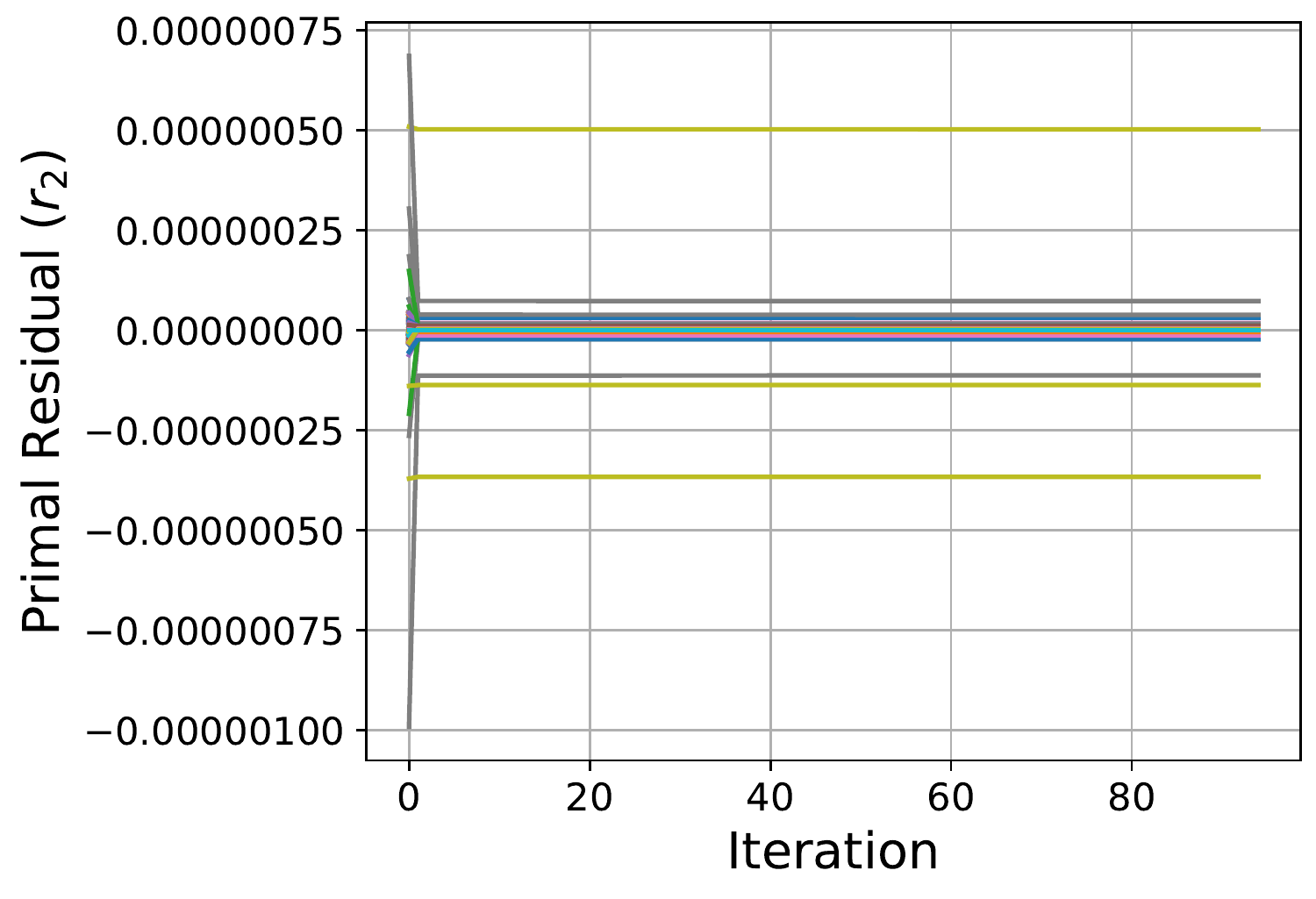}
		\caption{Primal residual $r_2$.}
		\label{F:Convergence_Residual2}
	\end{subfigure}

	\begin{subfigure}{0.325\linewidth}
		\includegraphics[width=\linewidth]{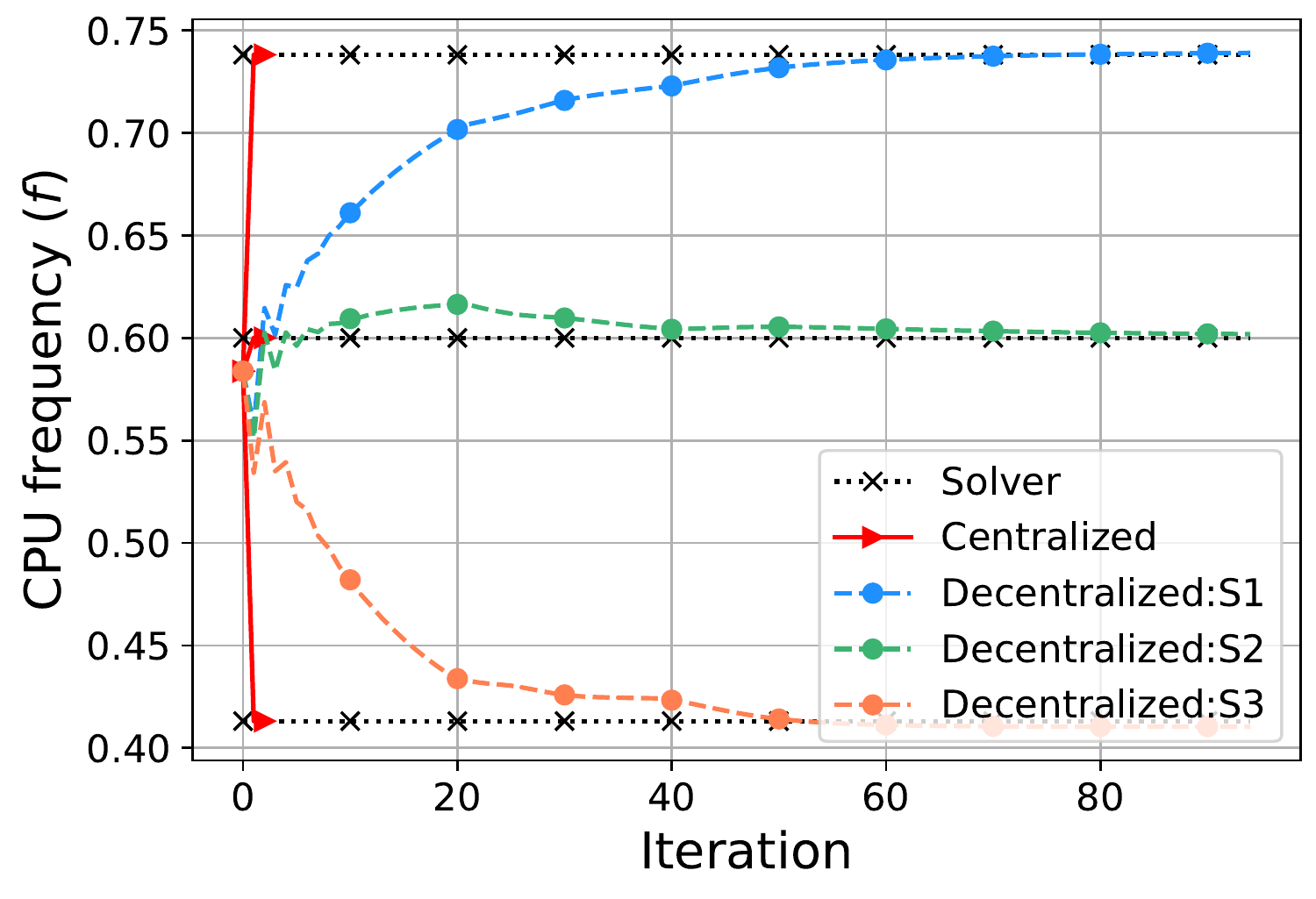}
		\caption{CPU allocation at UE $1$.}
		\label{F:Convergence_f}
	\end{subfigure}
	\begin{subfigure}{0.325\linewidth}
		\includegraphics[width=\linewidth]{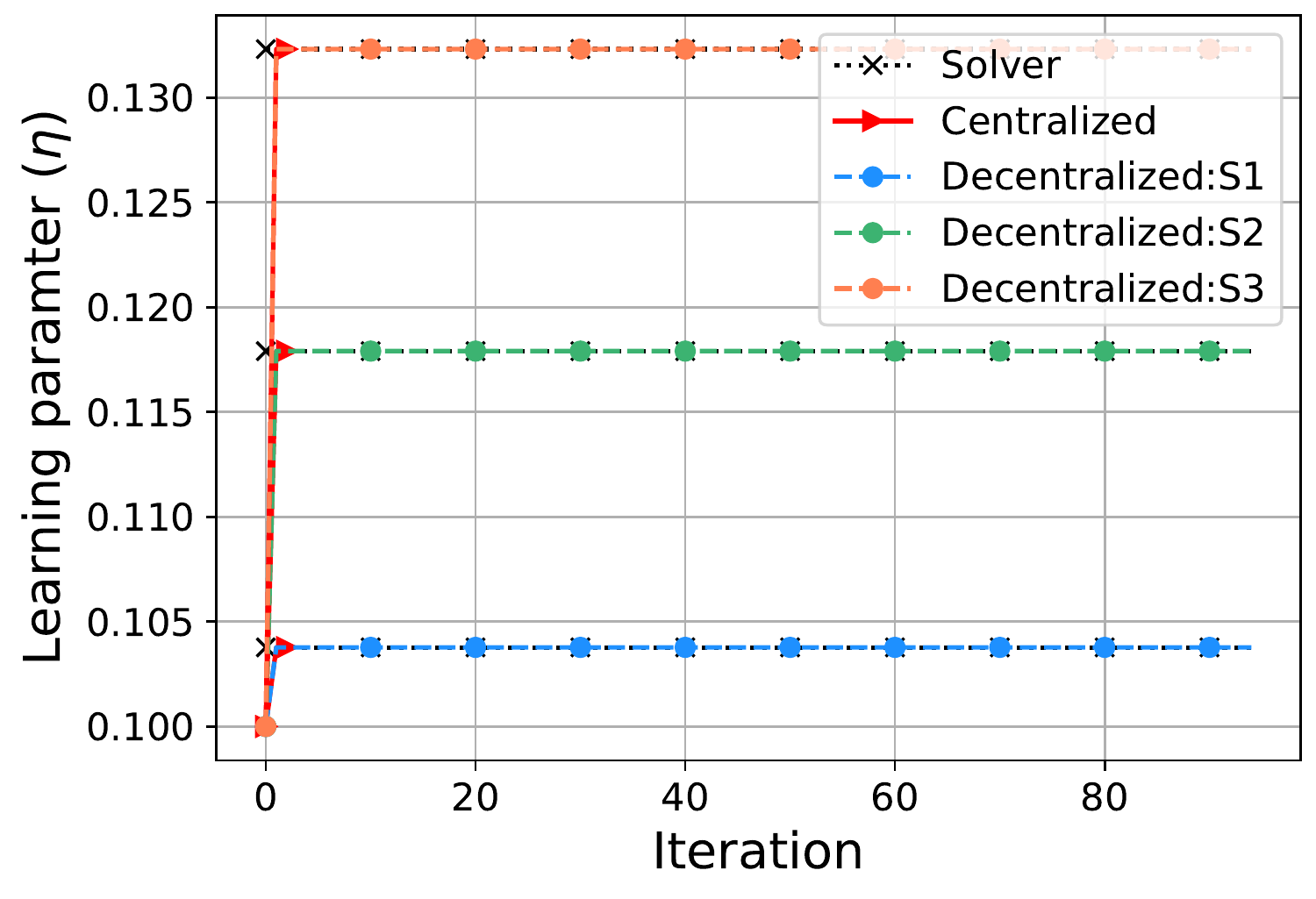}
		\caption{Hyper-learning rate for each service.}
		\label{F:Convergence_eta}
	\end{subfigure}
	\begin{subfigure}{0.325\linewidth}
		\includegraphics[width=\linewidth]{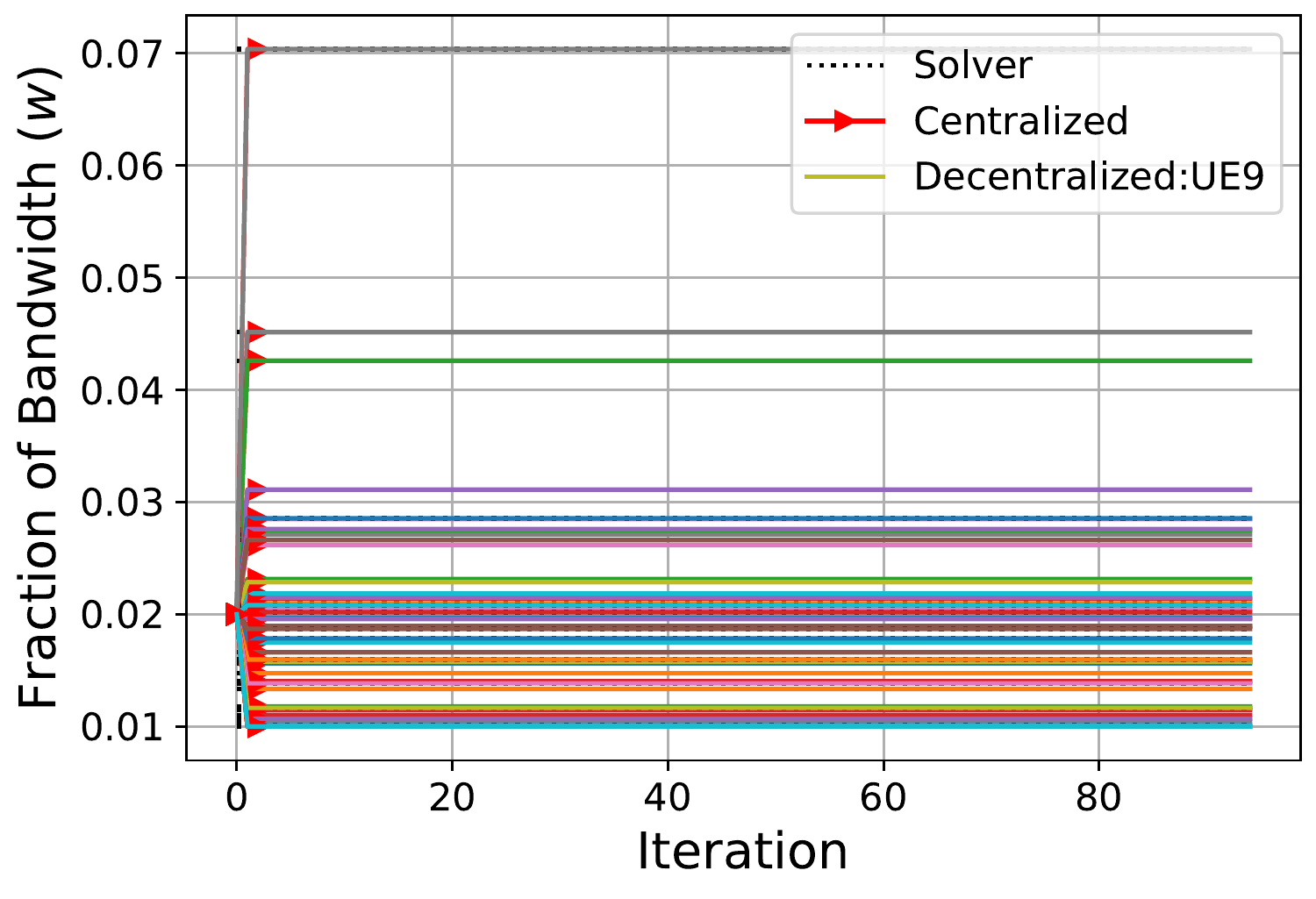}
		\caption{Bandwidth allocation for each UE.}
		\label{F:Convergence_w}
	\end{subfigure}
	\caption{Convergence experiment.}
	\label{F:convergence}
\end{figure*}
	
	Note that the chosen parameters $\rho_1$, $\rho_2$ in the augmented and proximal terms could affect the convergence performance of the decentralized approach. Furthermore, for a particular global convex problem with additional running conditions, JP-ADMM obtains $o(1/k)$ convergence rate according to Theorem 2.2 in \cite{jpadmm2017}, where $k$ denotes the number of iterations. Specifically, $\| x^k - x^{k+1} \|^2_{M_x} = o(1/k)$ where $x^k$ is the primal solution at the iteration $k$ and $M_x$ is defined in \cite{jpadmm2017}. Accordingly, the gaps between updated primal variables become gradually smaller throughout the iterative updates and the solutions converge toward the optimal ones. Conventionally, for a global convex problem, JP-ADMM converges faster than the dual decomposition method \cite{jpadmm2017} but still requires a higher number of iterations compared to Gauss-Seidel ADMM as shown in the simulation results of \cite{lin2015global}. For the multi-convex case, miADMM can guarantee the global convergence to the Nash point (i.e., stationary point) with the convergence rate $o(1/k)$ \cite{miADMM2019}.
	In the next section, we provide numerical results for the convergence performance and the efficiency of the proposed algorithms.

	\section{Performance Evaluation} \label{S:Sim}
	\subsection{Numerical Settings}
	In this work, we assume that three learning services are deployed at the edge networks. Moreover, 50 heterogeneous UEs are positioned within the coverage area of the base station to participate in the federated learning system. Similar to our prior works \cite{FL_TON2019,tranFederatedLearningWireless2019}, we consider that the channel gain between the base station and the UE follows the exponential distribution with the mean $g_0(d_0/d)^4$  where $g_0 = -40\,dB$, the reference distance $d_0=1\,m$ between BS and UEs. Here, the actual distance $d$ between the UEs and the base station is uniformly generated between $[2, 50]\, m$. 
	Furthermore, the uplink system bandwidth $B=20\,MHz$ is shared amongst UEs, the Noise power spectral is $10^{-10}\,W$, and the transmit power of UEs and BS are $10\, W$ and $40\, W$, respectively. 
	In this work, we assume that the size of the uploaded local model and downloaded global model is the same and it is set to $v_s \in\{100, 200, 300\}\,KB$ for each service.

	For the local training model at UEs, we first set the training data size of UEs in each learning service following a uniform distribution in $10-20\, MB$. 
	The maximum computation capacity (i.e., CPU frequency) at each UE is uniformly distributed between $[1, 2]\, GHz$. The required CPU cycles $c_s$ to train one bit of data at the UE for each learning service is $ \{50, 70, 90\} \, cycles$. Furthermore, the minimum required CPU frequency for each service is $f^{min}_s = 0.1\, GHz$.
	We consider that the effective capacitance coefficient is the same for all UEs as $\beta_n = 2 \cdot 10^{-28}$ and the trade-off parameter $\kappa_s$ is set to $0.2$. For the federated learning parameters, we set $L=1, \, \beta=0.5, \gamma=1,$ and $c=1$. Then, the relative accuracy of the local problem at UEs for each service is $\theta \in \{0.07,0.06,0.05\}$. This setting reflects the CPU frequency requirement and model size as above and defines the required number of local iterations for each learning service correspondingly. Finally, for the algorithm setting, we set the convergence thresholds in the algorithms as $\epsilon_1$ and $\epsilon_2$ are $10^{-4}$ and $ 10^{-5}$, respectively. Then, the values of the parameters $\rho_1$, $\rho_2$, and $\nu$ are $1000$, $10$, and $1500$ respectively.
	
		
		\begin{figure*}[t]
			\centering
			\begin{subfigure}{0.325\linewidth}
				\includegraphics[width=\linewidth]{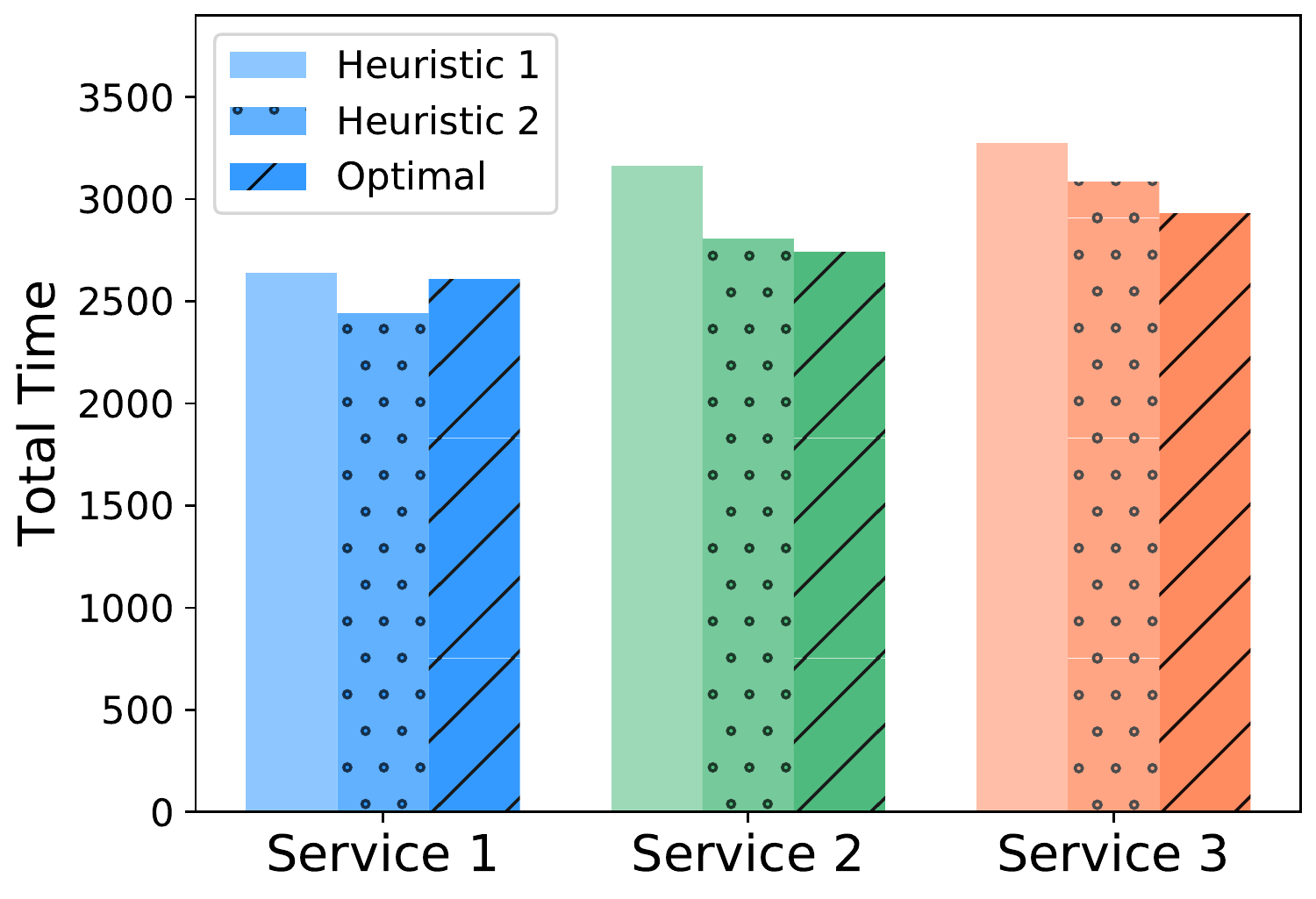}
				\caption{Time Comparison.}
				\label{F:T_Comparison}
			\end{subfigure}
			\begin{subfigure}{0.325\linewidth}
				\includegraphics[width=\linewidth]{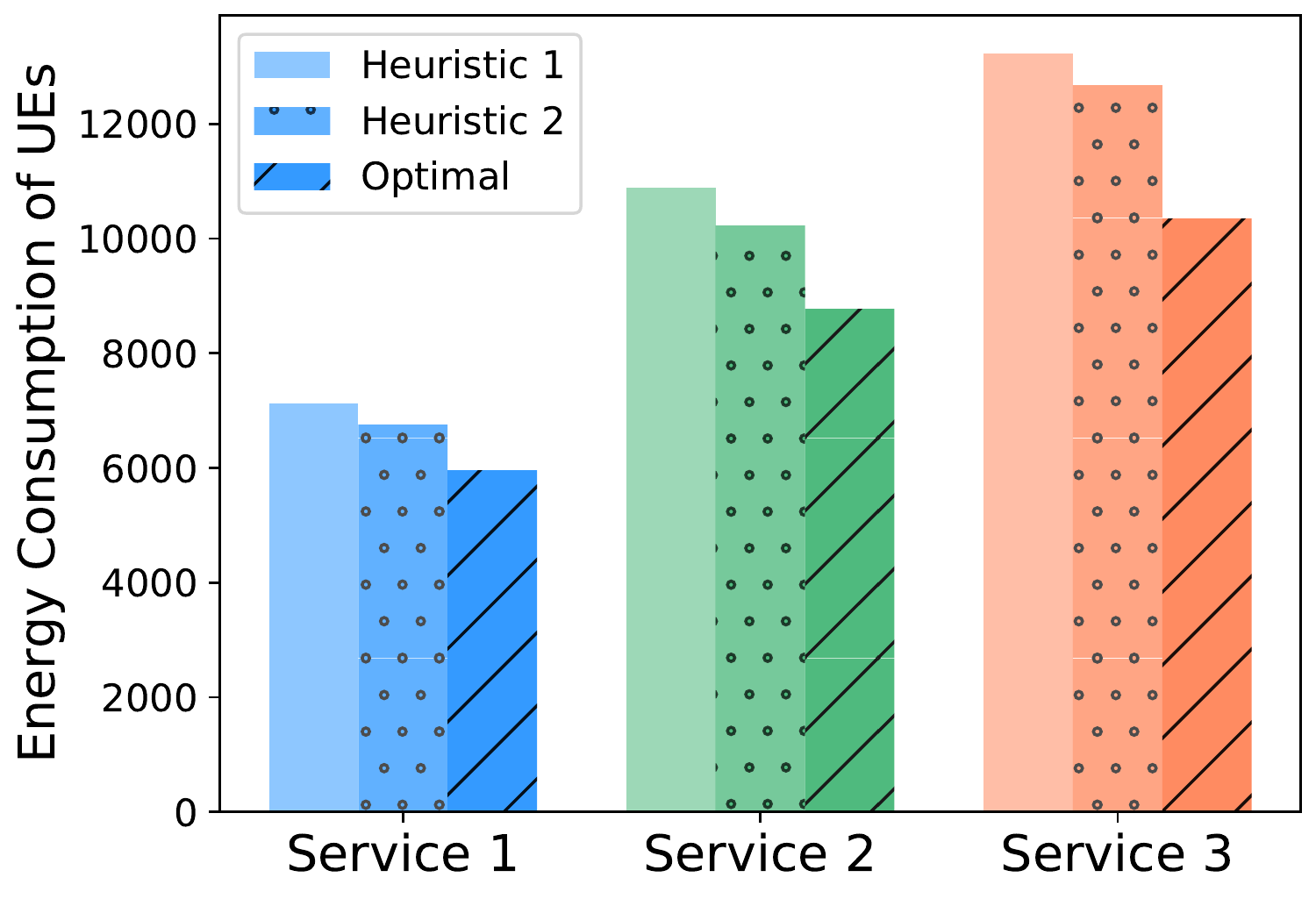}
				\caption{Energy Comparison.}
				\label{F:E_Comparison}
			\end{subfigure}
			\begin{subfigure}{0.325\linewidth}
				\includegraphics[width=\linewidth]{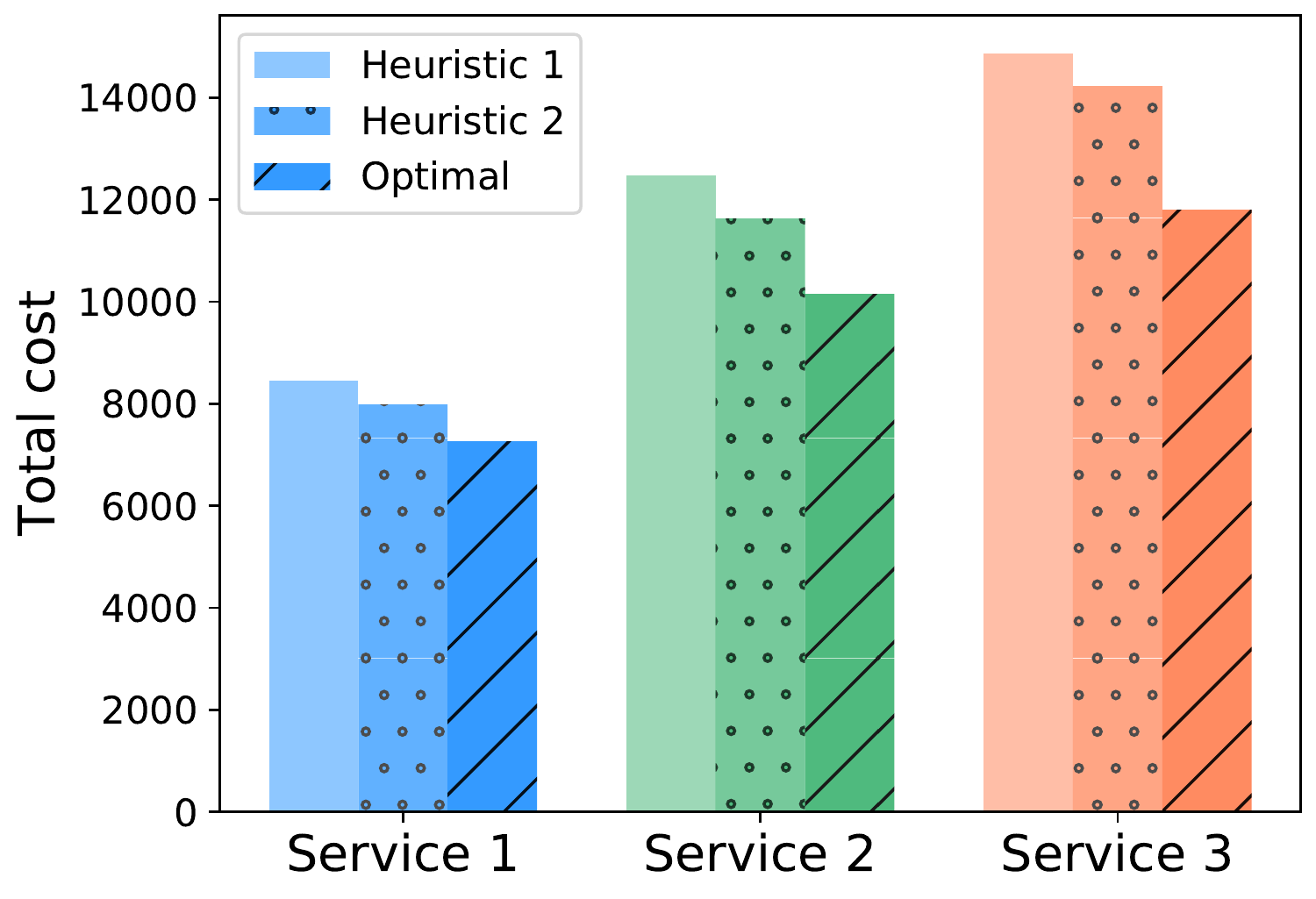}
				\caption{Service Cost Comparison.}
				\label{F:Service_Cost_Comparison}
			\end{subfigure}
			\caption{Cost Comparison.}
			\label{F:comparison}
		\end{figure*}
		
		\begin{figure*}[t]
			\centering
			\begin{subfigure}{0.325\linewidth}
				\includegraphics[width=\linewidth]{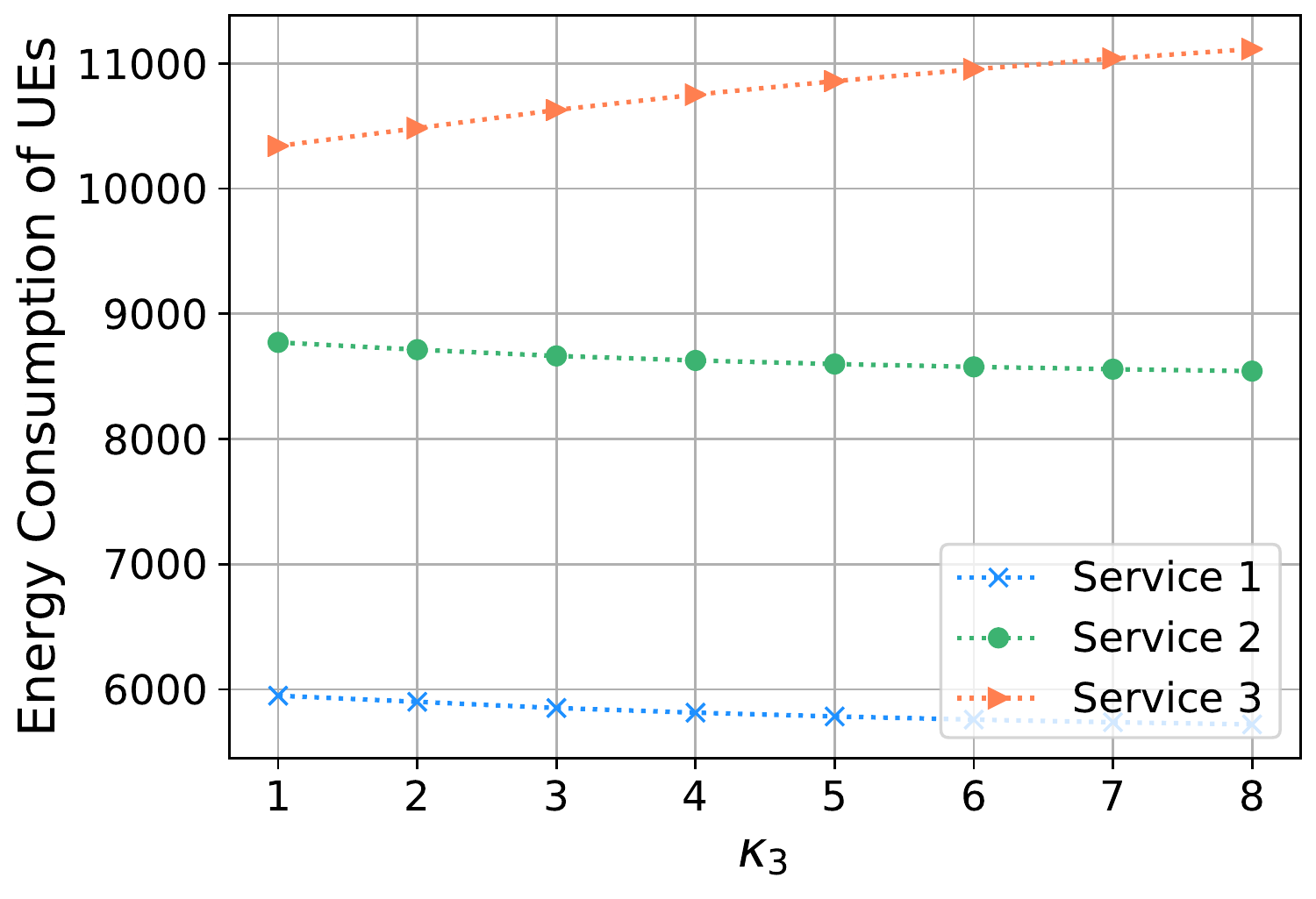}
				\caption{Energy consumption of services by increasing $\kappa_3$.}
				\label{F:Priority_Energy}
			\end{subfigure}
			\begin{subfigure}{0.325\linewidth}
				\includegraphics[width=\linewidth]{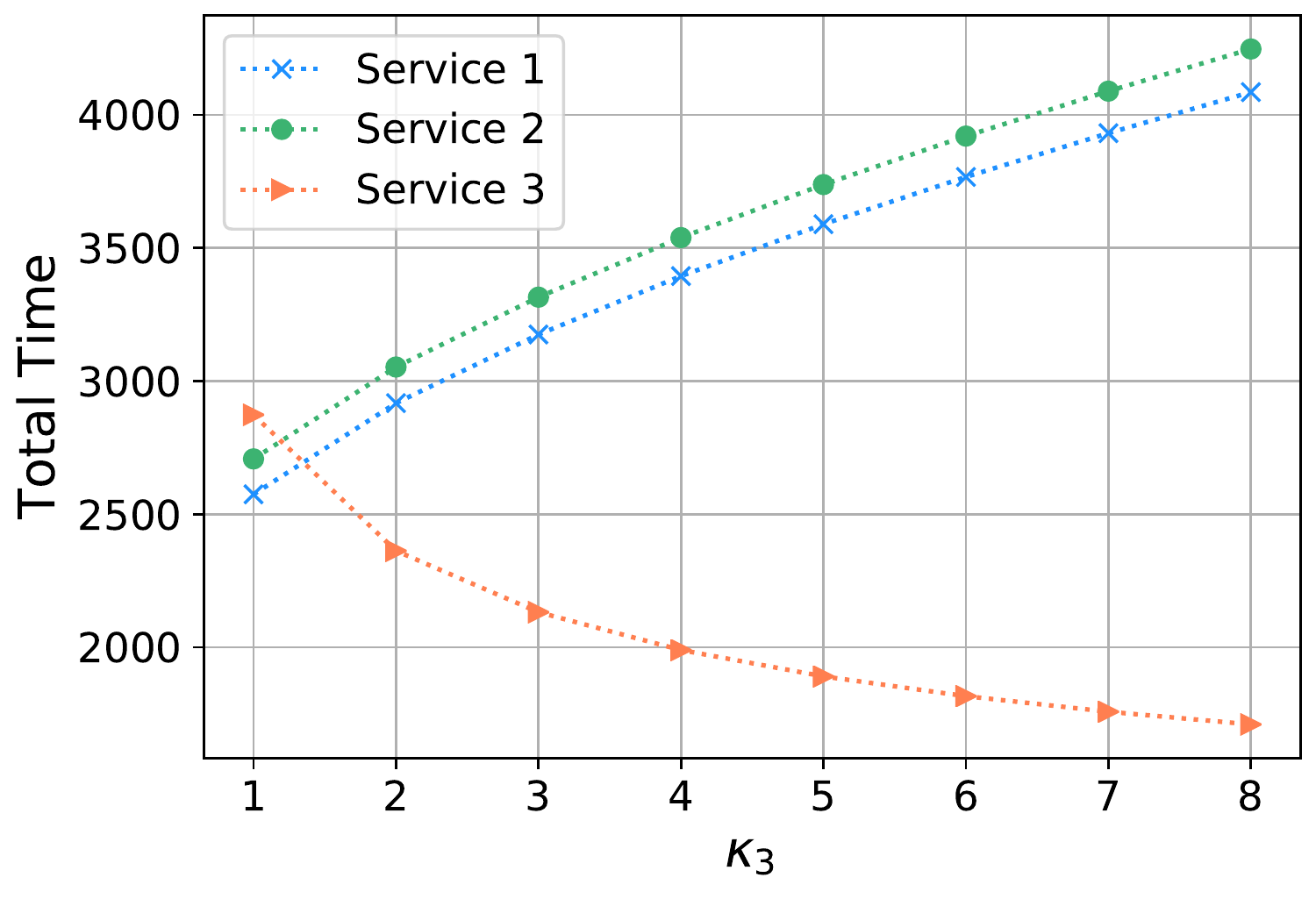}
				\caption{Total time of services by increasing $\kappa_3$.}
				\label{F:Priority_Time}
			\end{subfigure}
			\begin{subfigure}{0.325\linewidth}
				\includegraphics[width=\linewidth]{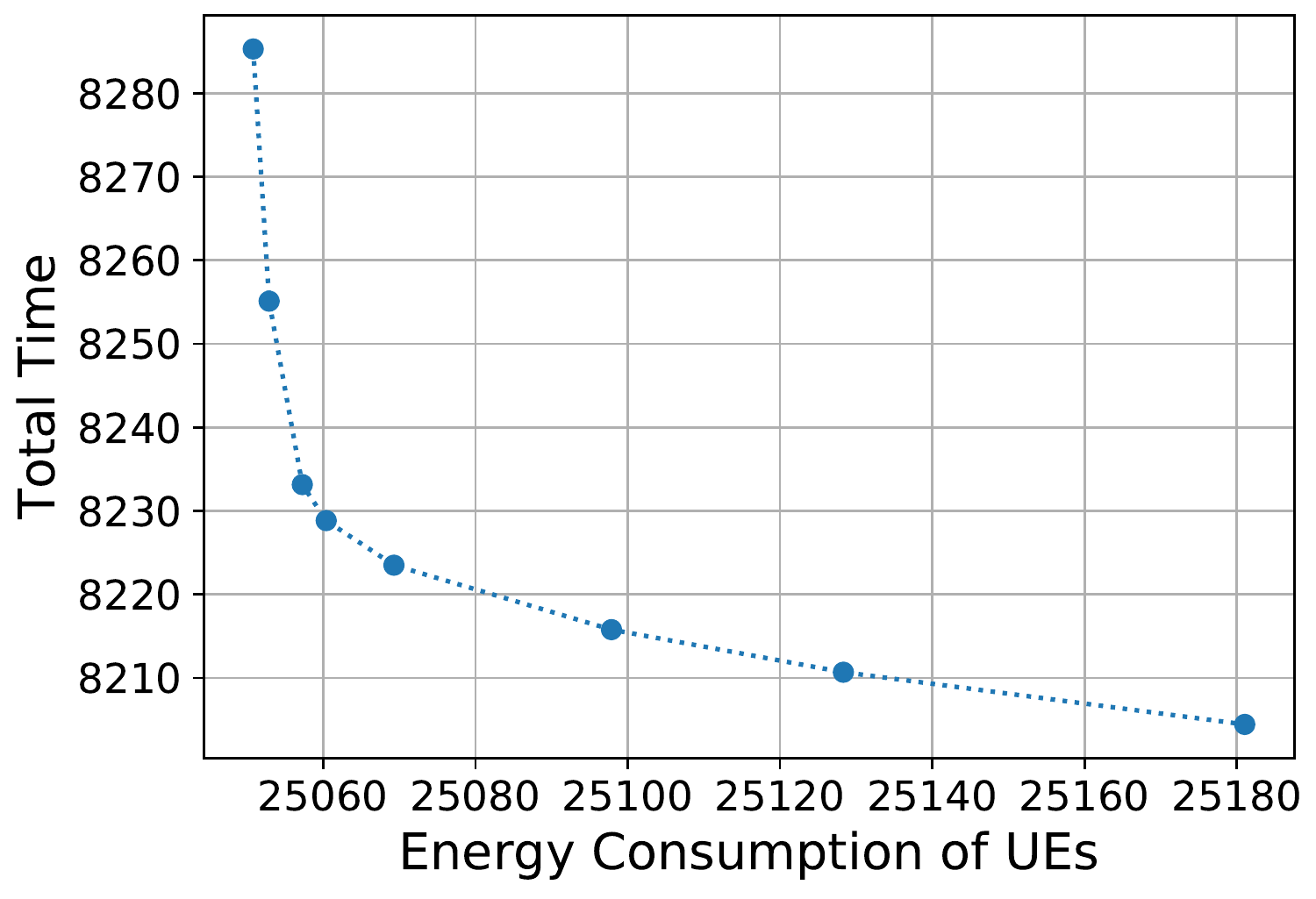}
				\caption{Trade-offs in objectives by reducing $\kappa_s$.}
				\label{F:Pareto}
			\end{subfigure}
			\caption{The trade-offs in energy consumption and total running time by varying $\kappa$.}
			\label{F:Trade-off}
		\end{figure*}
	
	\subsection{Numerical Results}
	We first illustrate a realization for the random location and local dataset size of UEs as shown in Fig. \ref{F:Distance_DataSize}.	
	For this realization, we demonstrate the convergence of the total cost, primal residual, CPU allocation, bandwidth from two solution approaches in Fig. \ref{F:convergence}.
	Accordingly, the centralized approach solely requires one iteration to get the optimal solution and the decentralized solution approach performs an iterative update process with many iterations to achieve that convergence condition such as the changes of solutions below small thresholds. Starting from the same initial points, the centralized approach is quickly converged within one iteration while the decentralized approach requires $95$ iterations to achieve the same solution in Fig. \ref{F:Objective}. Even though, the decentralized algorithm needs only 35 iterations to get almost similar cost compared to the optimal one, however, the allocated CPU frequency in Fig. \ref{F:Convergence_f} needs more iterations to obtain the same optimal solution from the solver or centralized algorithm. Different from a slow convergence of CPU frequency, the bandwidth solutions are quickly converged after 3 iterations and so the primal residual $r_2$ in the equation \eqref{E:pri_res2}. In practical usage, we can stop when the primal residual starts converging to zero after $55$ iterations as illustrated in Fig. \ref{F:Convergence_Residual1}, and \ref{F:Convergence_Residual2}. As a result, these solutions still guarantees to be a feasible solution and obtain such a very similar to optimal cost. In light of this observation, we later test the convergence performance of the integrated early stopping strategy in the decentralized algorithm.

	Now, we will discuss the characteristic of the optimal solution in Fig. \ref{F:convergence} as follows. As an example, by the reason of Service $1$ having the smallest local model parameter size to update, it has the highest hyper-learning rate and correspondingly takes more global rounds, less number of local iterations. Thereby, in order to proceed with the local learning and perform more global rounds quickly, Service 1 occupies most of the CPU frequency of UEs. Unlike Service 1, Service 3 has the lowest hyper-learning rate, takes the least CPU frequency, and performs less global rounds. 
	As the learning scheme in this work follows a synchronous federated learning, all UEs have to complete the local update and send the local weight parameters to the server before updating model at the MEC server. 
	Thus, the users who are far from the base station and have more training data to process will require longer time to train, then upload the local model. Accordingly, these devices receive the larger fraction of the uplink system bandwidth to upload their local parameters to the server. As shown in Fig. \ref{F:Distance_DataSize}, UE48 is one of the furthest UEs from BS and has a larger amount of local data. Therefore, the largest fraction of the uplink system bandwidth is allocated to UE48 as shown in Fig. \ref{F:Convergence_w}. 
	
	Moreover, we compare the optimal learning time, energy consumption, and total cost with two heuristic approaches as shown in Fig. \ref{F:comparison}. 
	In the first heuristic approach, the CPU frequency of the UEs is equally allocated to the learning services. Moreover, the uplink system bandwidth is equally allocated to the UEs as well to upload the local learning weight parameter to the server. Then, FLO decides solely the optimal hyper-learning rate for each learning service by solving the \SubLearningA problem. The second heuristic approach is adopted to proportionally allocate the local CPU based on the local data size (i.e., $D_{s,n}$) of each service at UEs and allocate bandwidth based on the transmission capacity of each UE.
	From the figures, we observe that the cost including the learning time and energy consumption of UEs for all services is reduced more than $18\%$ and $16\%$ than that of the Heuristic $1$ and Heuristic $2$ strategies.
	Among three learning services, Service 1 has the lowest CPU requirement and the smallest size of local model parameters. Therefore, it needs the lowest learning time and energy consumption as well.
	However, the minimum local performance at UEs of Service 1 compels more global rounds and the lowest values of the hyper-learning rate $\eta$ than that of Service 2, and 3.

	Furthermore, we vary the trade-off parameter to study the effects of the conflicting goals of minimizing the time cost and energy cost of each FL service. As increasing the trade-off parameter (i.e., $\kappa_s$), the \Opt gives a higher priority to minimize the running time than that of the energy consumption of UEs. Consequently, Fig. \ref{F:Pareto} shows the decreasing curve of running time. Meanwhile, the services require more energy consumption from UEs. Moreover, the priority of services in terms of running time can be controlled by varying the trade-off parameter $\kappa_s$ of the service $s$. Accordingly, the service which has a higher priority can be set with a higher value of $\kappa_s$ to reduce the learning time in solving the \Opt problem. In Fig. \ref{F:Trade-off}, we only increase the trade-off parameter $\kappa_3$ of Service 3 to demonstrate the higher priority of Service 3 than that of the other services. As a result, the total running time of Service 3 can be decreased while the total time of Service 1 and 2 are increased.  On the other hand, the energy consumption of UEs to serve Service 3 is increased while the energy consumption of UEs to serve other services is slightly decreased.
	
\begin{figure}[t]
		\includegraphics[width=\linewidth]{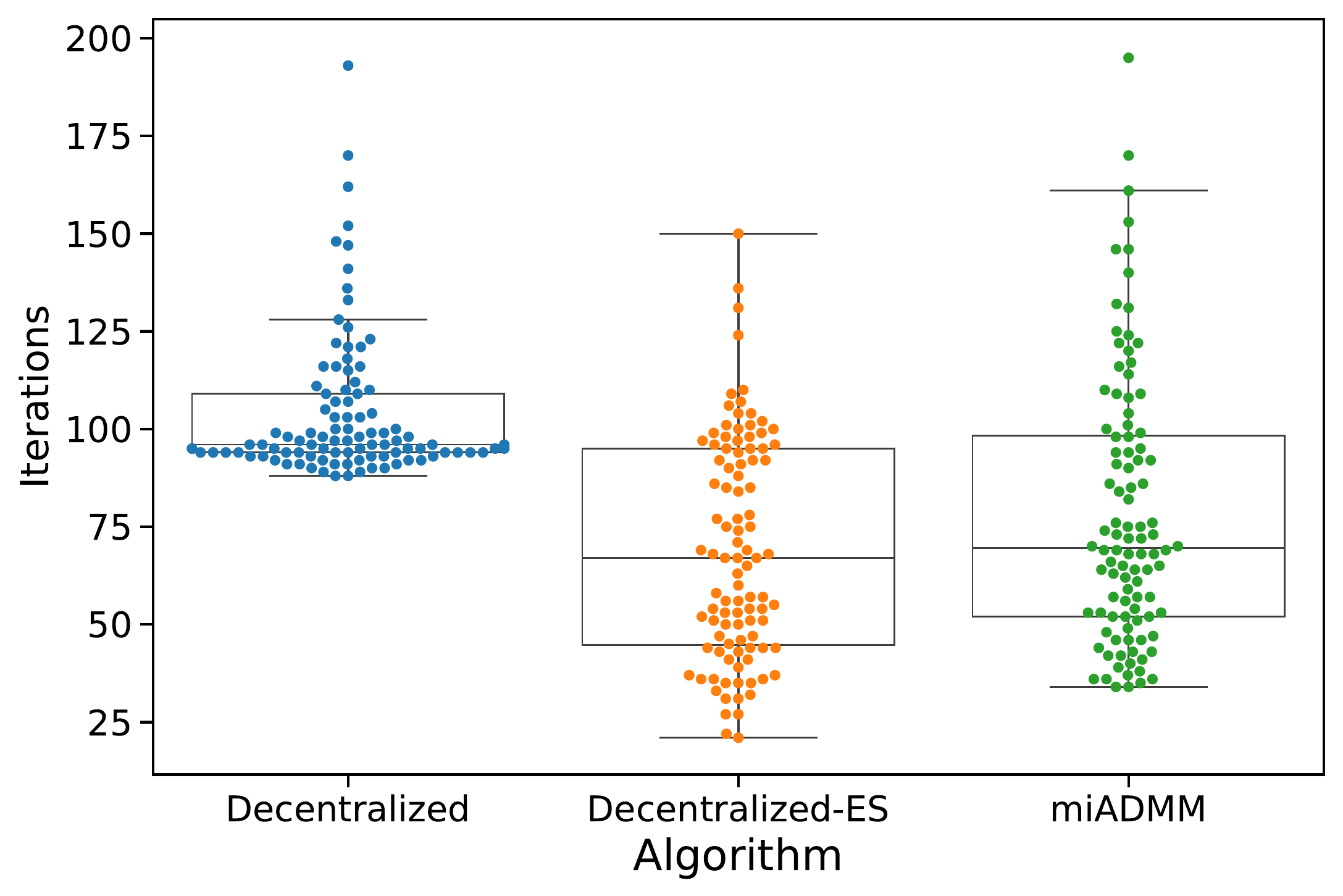}
		\caption{Convergence performance of the decentralized algorithms in 100 realizations.}
		\label{F:Iters_Stat_100}
	\end{figure}

	To boost the convergence speed, we can apply the early stopping strategy for the decentralized algorithm, namely the Decentralized-ES algorithm, by using the convergence condition on the primal residuals instead of the primal variables as we discuss above. In Fig. \ref{F:Iters_Stat_100}, we run $100$ realizations for the random location and local dataset size of UEs while keeping the other settings to validate the convergence speed of the decentralized approach using the Jacobi-Proximal scheme for multi-convex ADMM, the early stopping version of the decentralized algorithm and the original multi-convex ADMM algorithm using Gauss-Seidel scheme. Accordingly, we observe that the median values of the required iterations of the decentralized, decentralized-ES and miADMM algorithms for convergence are $96$ iterations, $67$ iterations, and $70$ iterations, respectively. The decentralized algorithm requires higher number of iterations on average than the original miAdMM algorithm. However, the early stopping strategy helps to speed up the convergence rate and obtain nearly the optimal cost. Statistically, there is only a $0.05\%$ difference with the optimal cost. Note that, the original miADMM does not fully support the parallel operation in the primal problem and requires cyclic learning service operating based on Gauss-Seidel update scheme. Besides, even though the proposed decentralized algorithms require a higher number of iterations to convergence, they provide privacy preserved and flexible approaches to independently control the learning process and resource allocation for each learning service. 
	Note that, if the stringent time-constrained is required, the centralized algorithm is applicable because it can provide the solution within two iterations.

	\section{Conclusion} \label{S:Concls}
	In this paper, we analyzed a multi-service federated learning scheme that is managed by a federated learning orchestrator to provide the optimal computation, communication resources and control the learning process. We first formulate the optimization model for computation, communication resource allocation, and the hyper-learning rate decision among learning services regarding the learning time and energy consumption of UEs. We then decompose the proposed multi-convex problem into three convex sub-problems and solve them alternatively by using the block coordinate descent algorithm in the centralized manner. Besides, we develop a decentralized algorithm to preserve the privacy of each learning service without revealing the learning service information (i.e., dataset information, exchange local updates information between UEs and the MEC server, the number of CPU cycles for each UE to execute one sample of data) to FLO. The simulation results demonstrate the superior convergence performance of the centralized algorithm and the efficiency of the proposed approach compared to the heuristic strategy. Furthermore, by experiment, the early stopping strategy can boost the convergence speed of the decentralized algorithm with an infinitesimal higher value than the optimal solution.
	
	To scale up our design, different MEC servers in different cells can independently allocate their resources. The recent works have analyzed the hierarchical federated learning framework in \cite{abad2020hierarchical, liu2020client} and game-strategic formulation that provide promising directions to extend this work regarding the scalability issue. We advocate the wireless dynamic and packet losses impacts as in \cite{chen2019joint} on the bounds of global iterations in the \FEDL algorithm and \Opt problem are also important to employ FL scheme to the realistic communication scenarios. We leave the possibly extensive analysis of the proposed approaches for our future works.
	
	\bibliographystyle{IEEEtran}
	\bibliography{On-Device-AI,minh}
	
	
	\newpage
	\clearpage
	
	\appendix
	\subsection{Proof for Lemma \ref{L:3}}
	
		In this proof, we skip the subscription $s$ for each service.
		
		According to the equation \eqref{E:K_Theta}, we have $ K_g(\Theta) = \frac{A_1}{\Theta}$ and $\Theta = \frac{2\rho\bigP{B\eta^2 + 1}}{C\eta -D\eta^2}$
		where
		\begin{align}
		A \defeq& \log \frac{F(w^0) - F(w^*)}{\epsilon} >0, \nonumber\\
		B \defeq& (1+\theta)^2\rho^2, \nonumber\\
		C \defeq& 2(\theta-1)^2- 2(\theta+1)\theta\rho^2, \nonumber\\
		D \defeq& \rho^2(\theta+1) (3\theta+1),  \nonumber 
		\end{align}
		and $\theta \in (0,1)$ is defined in \eqref{E:theta_approximation}, $\rho = \frac{L}{\beta}>1,\, L, \, \beta$ are defined in Assumption 1.
		
		Given $ 0 < \Theta < 1$ and $0 < \eta$,
		
		We have $C\eta -D\eta^2 >0  \Rightarrow C >0, \eta<\frac{C}{D}.$
		
		Thus, $B, C, D >0$
		
		We first verify the convexity of the \SubLearningB problem then derive a closed-form solution for this problem.
		The problem is strictly convex by validating the first and second derivatives of the objective function as follows
		\begin{align}
		g(\eta) &\defeq \frac{\Cal{C}(\eta)} {2\rho A_1\hat{\cal{C}}_s}  = \frac{B\eta^2 + 1}{C \eta -D\eta^2} \nonumber\\
		\Rightarrow \nabla 	g(\eta)
		 &= \frac{2B\eta (C \eta -D\eta^2) - (B\eta^2 + 1) (C- 2D\eta) }{(C \eta -D\eta^2)^2}, \nonumber\\
		 &= \frac{BC\eta^2+ 2D\eta  -C  }{(C \eta -D\eta^2)^2}.
		\end{align}
		\begin{align}
		\Rightarrow
		&\nabla^2 g(\eta) \nonumber\\
		&= \frac{2BC\eta+ 2D }{(C \eta -D\eta^2)^2} -2\frac{(BC\eta^2+ 2D\eta  -C)(C- 2D\eta) }{(C \eta -D\eta^2)^3} ; \nonumber\\
		&= \frac{2BC\eta+ 2D }{(C \eta -D\eta^2)^2} +2\frac{BC\eta^2 (2D\eta  -C) + (2D\eta - C)^2 }{(C \eta -D\eta^2)^3} ;\nonumber\\
		&= \frac{2D}{(C \eta -D\eta^2)^2} + 2\frac{BCD\eta^3 + (2D\eta - C)^2 }{(C \eta -D\eta^2)^3} >0. 
		\end{align}
		
		Thus, the problem is strictly convex and has a unique solution.
		
		\begin{align} 
		&\nabla \Cal{C} (\eta) =0  \nonumber\\
		&\Rightarrow BC\eta^2+ 2D\eta  -C  = 0 \nonumber\\
		&\Rightarrow (\text{Since } \eta>0), \eta^* = \frac{-D + \sqrt{D^2 +BC^2 }}{BC}.  \label{E:Sol_Sub1}
		\end{align}	
		
		Note that, in our prior work, the sufficient small value of $\theta$ can guarantee the feasibility of the problem and the closed-form solution \eqref{E:Sol_Sub1}. Also, This happens when many local iterations are required to fit the local models and causes the small values of $\theta$.

	\subsection{Convexity Proof for \SubCPUA problem}
	
	Firstly let us introduce the Lagrangian function of the \SubCPUA problem as follows
	\begin{align}
	&\Cal{L}(T^{\textrm{cmp}}, f, \lambda, \mu, \beta) = \sum_{s\in \mathcal{S}} \sum_{n \in \mathcal{N}} \lambda_{s,n} \left( \frac{c_sD_{s,n}}{f_{s,n}} - T_s^{\textrm{cmp}} \right) \nonumber\\
	&+ \sum_{s \in \mathcal{S}}  K_{l,s}K_g(\Theta^*_s)\left(\sum_{n \in \mathcal{N}} \frac{\alpha_n}{2}c_sD_{s,n}f^2_{s,n} + \kappa_s T_s^{\textrm{cmp}}  \right)  \nonumber\\
	&+ \sum_{n \in \mathcal{N}} \mu_n \big( \sum_{n\in \mathcal{N}} f_{s,n} - f_n^{\textrm{tot}} \big) + \sum_{s \in \mathcal{S}}\sum_{n \in \mathcal{N}}\lambda_{s,n}(f_{s, min}-f_{s,n}),   \label{E:Langrange1}  
	\end{align}
	where $\lambda, \beta \geq 0,$ and $\mu$ are Lagrangian multipliers. The first-order partial derivative of the Lagrangian function in \eqref{E:Langrange1} is as follows	
	\begin{align}
	&\frac{\partial \Cal{L}}{\partial f_{s,n}} = K_{l,s}K_g(\Theta^*_s) \alpha_n c_sD_{s,n}f_{s,n}-   \frac{\lambda_{s,n}c_sD_{s,n}}{f^2_{s,n}} + \mu_n - \beta_{s,n} ;\nonumber\\
	&\frac{\partial \Cal{L}}{\partial T_s^{\textrm{cmp}}} = K_{l,s}K_g(\Theta^*_s)\kappa_s + \sum_{n \in \mathcal{N}} \lambda_{s,n}.\nonumber
	\end{align}
	Then, the second-order derivative as follows
	\begin{align}
	&\frac{\partial^2\Cal{L}}{\partial f^2_{s,n}} = K_{l,s}K_g(\Theta^*_s) \alpha_nc_sD_{s,n} + \frac{2\lambda_{s,n}c_sD_{s,n}}{f^3_{s,n}} > 0;\\
	&\frac{\partial^2\Cal{L}}{\partial (T_s^{\textrm{cmp}})^2} = 0. 
	\end{align}
	The Hessian matrix of the Lagrangian function is positive semi-definite. Therefore, we can conclude that the \SubCPUA problem is a convex problem.
	
	\subsection{Convexity Proof for \SubBWA problem}
	
	The Lagrangian function of the \SubBWA problem of BW allocation is as follows
	
	\begin{align}
	&\Cal{L}(T^{\textrm{com}}, w, \nu, \zeta,\delta)  =  \sum_{s \in \mathcal{S}} K_g(\Theta^*_s) \left(\sum_{n \in \mathcal{N}}p_n\tau_{s,n}^{ul}(w_{n})+ \kappa_s T_s^{com}\right) \nonumber \\
	&+\sum_{s \in \mathcal{S}} \sum_{n \in \mathcal{N}} \nu_{s,n} ( \tau_{s,n}^{ul}(w_{n})+\tau_{s,n}^{dl} - T_s^{com})  \nonumber\\
	&+\sum_{n \in \mathcal{N}} \zeta_n ( w_{min} - w_n) + \delta\sum_{n \in \mathcal{N}}(1-w_{n}),  \label{E:Langrange2}
	\end{align}	
	where $\zeta, \nu \geq 0,$ and $\delta$ are Lagrangian multipliers.
	Then, the first-order derivative of the Lagrangian function in \eqref{E:Langrange2} as follows
	\begin{align}
	&\frac{\partial \Cal{L} }{\partial w_n} = \frac{- K_g(\Theta^*_s) v_s}{w_n^2B^{ul}\log_2(1 + \frac{h_np_n}{N_0})} \left( p_n + \nu_{s,n}\right) - \zeta_n;    \nonumber  \\
	&\frac{\partial \Cal{L} }{\partial  T_s^{\textrm{com}}} = K_g(\Theta^*_s) \kappa_s  - \sum_{n \in \mathcal{N}} \nu_{s,n}
	\end{align}
	The second-order derivative of is as follows:
	\begin{align}
	&\frac{\partial^2\Cal{L}}{\partial w_n^2} = \frac{ 2 K_g(\Theta^*_s) v_s}{w_n^3B^{ul} \log_2(1 + \frac{h_np_n}{N_0})} \left( p_n + \nu_{s,n}\right) >0;    \\
	&\frac{\partial^2\Cal{L}}{\partial (T_s^{\textrm{com}})^2} = 0. 
	\end{align}
	Therefore, we can conclude that the \SubBWA problem is a convex problem.

\end{document}